%% file: main.tex
\documentclass{article}
\usepackage[preprint]{neurips_2026}
\usepackage[T1]{fontenc}
\usepackage[utf8]{inputenc}
\usepackage{amsmath,amssymb,amsthm}
\usepackage{booktabs,tabularx,array}
\usepackage{graphicx,subcaption}
\usepackage{algorithm,algpseudocode}
\usepackage{xcolor,microtype}
\usepackage{tikz}
\usetikzlibrary{arrows.meta,calc,shapes.geometric}
\DeclareMathSizes{9}{9}{6}{5}
\usepackage{url}
\usepackage[colorlinks=true,allcolors=blue]{hyperref}
\hypersetup{
  pdftitle={Constructing Budgeted KGW Evasion through Token-Event Evidence Replacement},
  pdfauthor={Hongliang Liu},
  pdfsubject={Budgeted token-event editing of controlled KGW watermarks}
}
\newtheorem{proposition}{Proposition}
\newtheorem{corollary}{Corollary}

\newcommand{\ind}{\mathbf{1}}

\graphicspath{{figures/}}

\title{Constructing Budgeted KGW Evasion through Token-Event Evidence Replacement}
\author{Hongliang Liu\\Independent Researcher\\\texttt{phunter.lau@gmail.com}}
\date{}

\begin{document}
\maketitle
\begin{abstract}
\input{sections/abstract}
\end{abstract}
\input{sections/introduction}
\input{sections/background}
\input{sections/related_work}
\input{sections/setting}
\input{sections/evidence_capacity}
\input{sections/construction}
\input{sections/experimental_design}
\input{sections/results}
\input{sections/discussion}
\input{sections/conclusion}

\bibliographystyle{plainnat}
\bibliography{references}
\clearpage
\appendix
\input{sections/appendix_proofs}
\input{sections/appendix_evidence}
\input{sections/appendix_reporting_controls}
\input{sections/appendix_historical}
\input{sections/appendix_stopping}
\input{sections/appendix_execution}
\end{document}

%% file: sections/abstract.tex
KGW watermarks bias sampling to accumulate keyed detection evidence. We construct budgeted evasion through token-event evidence replacement using ordinary phrase rewrites. Our construction jointly selects compositions of LLM-proposed alternatives from an independent model and measures global token-edit distance on complete outputs. At the largest attainable cost within the budget, exact coverage search maximizes removal of original scored token pairs. It uses a known tokenizer without the key, detector feedback or queries to the watermarked model. Exact accounting separates source evidence removed, evidence created and length effects, with initial strength setting the required reduction. Separate signed-event analysis found that mean suppression arose mainly from removing watermark-enriched source events. In controlled, occurrence-counted, one-token-context KGW tests on short passages, 39 of 82 qualified, initially detected sources yielded rewrites with terminal detector score $Z<4$ that passed blinded source-relative model review, at a mean actual token-edit cost of 16.0\% under a 20\% allowance. To examine whether this initial requirement can be inferred, a separate historical 95-passage study estimated initial detector score $Z_0$ from generation traces with correlation 0.793 and an in-sample association beyond length. This aggregate information did not establish improved allocation, and the separate no-trace stopping test did not reliably exploit cheaper reviewed crossings in the candidate bank. Comparisons did not establish an advantage over simpler selection and ordinary constrained rewriting. The construction makes budgeted evidence replacement executable while exposing limits in signed reduction and recognition of sufficient editing.

%% file: sections/introduction.tex
\section{Introduction}
\label{sec:introduction}

LLM watermarking embeds statistical evidence during generation and permits
comparatively inexpensive detection
\citep{kirchenbauer2023watermark,dathathri2024scalable}. In soft KGW, a secret key
and token context select a green vocabulary subset. A positive logit bias favors
that subset during sampling; detection reconstructs membership and tests the
resulting excess with a normalized $Z$ score. The signal depends on tokenizer
IDs and their contexts, whose boundaries can differ from words and propositions.
This creates a constrained editing problem: can ordinary phrase replacements
reduce detection evidence while preserving the source's claims and changing
only a limited number of tokens?

An edit budget limits how much of this evidence an editor can replace.
In the one-token-context setting studied here, each scored event is a token
paired with its predecessor. A token substitution can affect its own event
and the successor's context. A phrase rewrite can change more
events through retokenization, but those token changes also consume the budget.
The signed effect remains hidden: removing green-enriched source events can
reduce evidence, while newly created events can reinforce or offset that gain.
Even a favorable change must overcome the source's initial distance from the
threshold. Meaning-preserving alternatives, structural coverage and sufficient
signed reduction therefore impose distinct requirements.

Robustness analysis and probability-guided paraphrasing already study editing
attacks \citep{zhao2024robust,krishna2023paraphrasing,cheng2025sira,hwang2026bira}.
Color-aware substitution also studies low-edit evasion using queries to the
watermarked model \citep{wu2024color}.
Our question concerns how ordinary local rewrites operate under verified
whole-output token budgets. We examine three possible limits: whether the available
rewrites replace enough source events, whether their hidden signs yield sufficient
reduction, and whether the editor can recognize how much editing is needed.
A finite bank of local rewrites makes these limits separately measurable.

Our executable construction generates ordinary phrase alternatives and
jointly selects complete compositions. It measures token-edit cost on each
complete output and selects the composition that removes the most source events
at the largest attainable cost within each allowance. Selection uses no
source-model probabilities, generation traces or detector feedback; an independent
LLM supplies the rewrite proposals. Joint evaluation accounts for token boundaries
and events shared across replacements. Initial strength determines whether the
resulting reduction is sufficient. Enumeration implements the confirmed policy;
a verified prefix-sharing dynamic program reproduces its selections on the
same materialized candidate bank.

The spending rule permits comparisons between selectors at the same actual cost;
finding the cheapest successful rewrite
requires a separate stopping decision. Original events are enriched by
watermarked generation, which motivates their removal as a structural objective.
The signs of the new events remain unknown, so this objective supplies no
guarantee of crossing the detector threshold.

Figure~\ref{fig:examples} illustrates how this construction changes accumulated
evidence through localized wording changes. One phrase changes from
``exposed grave misconduct'' to ``revealed serious misconduct.'' The complete
rewrite combines three such regions and passes blinded source-relative model
review. Its token-edit distance is 14 for a 72-token source, about $19.4\%$:
the intervention is local in the prose but uses nearly the full $20\%$ allowance.
The final score falls from $6.290$ to $3.973$, just below the threshold of four.
Some earlier prefixes remain above four, so the example establishes terminal
escape only. This retrospectively selected illustration was exceptional:
it was the only quality-qualified success with an initial score above $6$ at this allowance.

\input{figures/example_panels}

We evaluated whether the construction yields reviewed escapes across generated
passages and whether probability weighting improves its selections. In the
fixed evaluation, coverage
and a surprisal-weighted rule each produced 39/82 escapes that passed blinded
source-relative model review, among passages that passed the source-quality gate
and were initially detected. Mean actual cost was $15.96\%$ under a $20\%$
allowance. These outcomes establish feasible budgeted constructions in the
tested setting. Comparisons within the same candidate set and a matched-cost
ordinary-rewriting pilot did not demonstrate an additional method benefit.

The study makes three contributions. First, it supplies an executable construction
that combines independent local proposals under verified whole-output token budgets
with measured detector and quality outcomes. Second, signed-event accounting
connects that construction to initial headroom and token-budget limits, while
complete candidate enumeration separates missed choices from absent crossing
opportunities. Third, it examines estimation of the initial evidence requirement
and recognition of sufficient editing.
A historical study estimates initial detector score $Z_0$ from generation
observations without the key. Follow-up inference on the confirmation bank
tests whether estimated source evidence improves matched-cost selection. The bank
contains cheaper reviewed crossings that a separate no-trace stopping rule does
not reliably exploit; added allocation value from the historical estimate beyond
length remains unresolved. These contributions concern the original one-token-context KGW
setting; human fidelity and transfer require further validation.

%% file: figures/example_panels.tex
\begin{figure}[!tp]
  \centering
  \input{figures/figure_one_subfigures}
  \caption{Sparse rewriting changes the accumulation of detector evidence.
  (a) Exact original-CUDA prefix scores for the successful illustration and its
  frozen selected rewrite end above and below the terminal threshold,
  respectively. Each curve uses its own token-prefix length; equal counts
  need not align in content. Numbered gray windows group paired phrase
  regions; blue/orange strips and table ranges locate them on each token path.
  (b) Bold words mark wording changes within the larger selected spans, which
  include unchanged context. These locations are not independent causal
  score contributions. The output has some prefixes above four, so this
  example establishes terminal escape only. The retrospectively selected
  output had global token-edit distance 14 for a 72-token source, yielded 70 output tokens, and passed
  relative and polished model review. Coverage and weighting selected the same
  output, reproduced in Appendix~\ref{app:complete-example}.}
  \label{fig:examples}
\end{figure}

%% file: figures/figure_one_subfigures.tex
\begingroup
\input{figures/figure_one_styles}
\captionsetup[subfigure]{font=small,labelfont=bf,justification=centering,skip=3pt}
\begin{subfigure}{\linewidth}
  \centering
  \input{figures/figure_one_trajectory}
  \caption{Localized phrase changes reduce accumulated evidence; the final score crosses the threshold.}
  \label{fig:examples-trajectory}
\end{subfigure}
\par\medskip
\begin{subfigure}{\linewidth}
  \centering
  \input{figures/evasion_replacement_table}
  \caption{Exact phrase replacements linked to the numbered regions above.}
  \label{fig:examples-phrases}
\end{subfigure}
\endgroup

%% file: sections/background.tex
\section{Background and related work}
\label{sec:background}

\paragraph{Embedding and detection.}
A generative text watermark couples output choices to a designated statistical signal, which a corresponding detector tests in a candidate passage. Research balances detectability, language quality, computational cost, and editing robustness. KGW uses context-keyed vocabulary partitions \citep{kirchenbauer2023watermark}, while Unigram-Watermark fixes the vocabulary grouping \citep{zhao2024robust}. Distortion-free constructions couple sampling to keyed randomness while preserving the text distribution up to a specified generation budget \citep{kuditipudi2023distortionfree}. SynthID-Text provides production-oriented context for sampling-time embedding and efficient detection \citep{dathathri2024scalable}. These families have different event representations and scoring rules. A positive test concerns the watermark-associated process under its assumptions; it certifies neither factual correctness nor personal authorship.

\paragraph{The KGW signal.}
Soft KGW promotes a context-selected green subset through a positive logit bias and detects excess green events relative to nominal fraction $\gamma$ \citep{kirchenbauer2023watermark}. Membership is defined on tokenizer IDs, whose boundaries can differ from words and propositions. Our setting uses ordered predecessor/current-token pairs with occurrence counting. Repeated identities share a keyed state, which affects statistical dependence. Section~\ref{sec:setting} defines the exact statistic and information boundary; the analysis and experiments concern that historical adjacent-token configuration.

%% file: sections/related_work.tex
\label{sec:related}

\paragraph{Editing robustness.}
The edited-token and successor-context effect appears in KGW, and subsequent
work provides editing-robustness analysis
\citep{kirchenbauer2023watermark,zhao2024robust}. Character-level perturbation
work explicitly analyzes attack range through retokenization and studies both
zero-feedback corruption and reference-detector-guided genetic search
\citep{zhang2026character}. Our support statements specialize local counting
with occurrence-multiset cancellation, length change and signed margin units;
the constructor admits ordinary phrase replacements. Shared counting conventions
are required for constant comparisons. Cryptographic constructions also provide
edit-distance robustness under computational and alphabet-size assumptions
\citep{golowich2024edit}. Our historical KGW result does not extend to those
constructions. The experiments use a pinned MarkLLM implementation
\citep{pan2024markllm}.

Reliability experiments study human and model paraphrasing, surviving source
fragments and detection of watermarked spans within longer mixed documents
\citep{kirchenbauer2024reliability}. Their longer-text and localized-detection
findings delimit the inference from a short-source terminal crossing in our
controlled setting. Surviving evidence can remain detectable under another
observation length or detector policy.

\paragraph{Rewriting and selection information.}
DIPPER studies contextual paragraph paraphrasing
\citep{krishna2023paraphrasing}. SIRA targets tokens through self-information
\citep{cheng2025sira}, BIRA applies negative logit bias to a surprisal-derived
suppression set \citep{hwang2026bira}, and Smoothing uses target-API probabilities
to identify low-confidence positions for reference-model resampling
\citep{chang2025smoothing}. These studies
establish probability-guided rewriting as prior work. Our final selector uses
token-event coverage without source-model probabilities; the historical
trace-assisted initial-strength estimate is a separate diagnostic.

\paragraph{Methodological comparison with SCTS.}
Low-edit evasion has direct precedent in Self Color Testing-based Substitution
(SCTS), which also operates without the key. It queries the watermarked model
to infer relative colors from sampled
frequencies, then scans for substitutions inferred to change green to red.
Its sequential rule uses a successor-context heuristic and evaluates normalized
word-edit budgets \citep{wu2024color}. Our constructor makes no such queries:
it enumerates complete compositions of independently proposed phrases and verifies
global token cost before maximizing removed-event coverage at the selected dose.
This comparison separates the information used for selection, the search rule
and the unit of cost; it supplies no matched performance ranking.

Historical initial-strength inference estimates the required reduction;
additive-utility DP optimizes supplied candidate effects under the companion's
state assumptions. It does not reveal their hidden signs or certify crossing.
The confirmed constructor uses neither this DP nor the initial-score estimate.
The newly verified coverage DP implements its structural policy
(Section~\ref{sec:construction}); sufficient-dose recognition remains a
separate question.

Watermark stealing learns approximate watermark rules from
queried outputs \citep{jovanovic2024stealing}; De-mark probes watermark strength
and color assignments through model queries \citep{chen2025demark}.
Published rates under different
access, cost units and fidelity criteria do not establish a matched advantage.

\paragraph{Quality-constrained removal.}
Watermarks in the Sand derives removal under quality-oracle and efficiently
mixing perturbation assumptions \citep{zhang2024sand}. Sandcastles in the Storm
examines practical mixing and quality-oracle failures, including disagreements
with human review \citep{harelcanada2025sandcastles}. These findings motivate
explicitly separating detector crossing from reviewed source preservation.
Our model-only judgments and finite local bank establish neither human fidelity
nor the assumptions required for unrestricted removal.

%% file: sections/setting.tex
\section{Setting and concrete interventions}
\label{sec:setting}

The editor receives source prose, a pinned tokenizer and the controlled watermark
family. It measures token changes and event incidence without querying the
watermarked generator or detector. It does not access the generator's weights,
key, colors, detector scores or original generation trace. Public-model
probabilities are available only to the labeled comparator.
Appendix~\ref{app:reporting-controls} records the complete access table.

Let $x=(x_1,\ldots,x_N)$ and $x'=(x'_1,\ldots,x'_{N'})$ be the canonical
tokenizations of the complete source and output. Actual edit cost is
\begin{equation}
q=d_{\mathrm{Lev}}(x,x'),\qquad T=N-1,\quad T'=N'-1,\quad
\ell=T'-T=N'-N.
\label{eq:math-cost}
\end{equation}
Insertions, deletions and substitutions each cost one. Edited words and
source-character sites can differ from this minimum distance because
retokenization and global alignment affect the complete output.

In the original context-width-one KGW configuration, the fixed key assigns each adjacent-token
identity $e=(a,b)$ a state $g_K(e)\in\{0,1\}$. Repeated identities share their
state. With occurrence count $n_x(e)$, centered state $y_K(e)=g_K(e)-\gamma$
and $\sigma=\sqrt{\gamma(1-\gamma)}$, the detector computes
\begin{align}
G_K(x)&=\sum_e n_x(e)g_K(e), &
A_K(x)&=G_K(x)-\gamma T=\sum_e n_x(e)y_K(e),\nonumber\\
Z_K(x)&=\frac{A_K(x)}{\sigma\sqrt T}, &
M_K(x)&=A_K(x)-\tau\sigma\sqrt T
       =\sigma\sqrt T\bigl(Z_K(x)-\tau\bigr).
\label{eq:math-detector}
\end{align}
The first token supplies context; scores require $T,T'>0$. We use
$\gamma=1/2$, threshold $\tau=4$ and the original CUDA partition implementation.
Repeated identities induce dependence, so this occurrence-counted $Z$ does
not itself establish an independent-event null calibration.

Reductions are source minus output: $\Delta Z=Z_K(x)-Z_K(x')$ and
$\Delta M=M_K(x)-M_K(x')$. An initially positive source has $Z_K(x)>\tau$;
strict escape requires $Z_K(x')<\tau$. Equality is excluded even though the
implementation's positive rule is $Z>\tau$. Quality-qualified escape
additionally requires the blinded source-relative review in
Section~\ref{sec:design}. Review follows sealed selection and supplies no
editing feedback. Human fidelity and external factual grounding remain
separate outcomes.

%% file: sections/evidence_capacity.tex
\section{Evidence replacement and crossing capacity}
\label{sec:evidence}

\subsection{Signed evidence determines the effect of a rewrite}

The schematic below illustrates how a replacement changes adjacent-token
events. Its colors represent hypothetical evaluator states, unavailable to
the editor; they are not inferred states for the measured example.
\begin{center}
\input{figures/figure_one_path}
\end{center}

A rewrite changes event multiplicities. Write $a_e=n_x(e)-n_{x'}(e)$ and
cancel common occurrences, using $[a]_+=\max(a,0)$, to define
\begin{equation}
d_e=[a_e]_+,\qquad c_e=[-a_e]_+,\qquad
D=\sum_e d_e,\qquad C=\sum_e c_e,\qquad C-D=\ell.
\label{eq:math-incidence}
\end{equation}
These are net occurrence counts: removing an identity and recreating it
elsewhere cancels in the terminal statistic. The corresponding evidence is
\begin{equation}
A_D=\sum_e d_e y_K(e),\qquad A_C=\sum_e c_e y_K(e),\qquad
L=A_D-A_C=A_K(x)-A_K(x').
\label{eq:math-signed-evidence}
\end{equation}
Substitution into Eq.~\eqref{eq:math-detector} gives
\begin{align}
\Delta M&=L+\Delta b,
&\Delta b&=\tau\sigma\bigl(\sqrt{T'}-\sqrt T\bigr),
\label{eq:math-margin-identity}\\
\Delta Z&=\frac{L}{\sigma\sqrt{T'}}+
Z_K(x)\left(1-\sqrt{\frac{T}{T'}}\right).
\label{eq:math-score-identity}
\end{align}
These identities require no randomness or independence assumptions. At unchanged
length, $\Delta M=L$ and $\Delta Z=L/(\sigma\sqrt T)$. For initial margin
$M_0=M_K(x)$, strict escape occurs exactly when
\begin{equation}
L>M_0-\Delta b,\qquad\text{equivalently}\qquad \Delta M>M_0.
\label{eq:math-crossing}
\end{equation}
Thus the same reduction can escape one source and leave a higher-headroom
source detectable. The tokenizer determines which events change; their keyed
states determine whether the created evidence offsets the removed evidence.

\subsection{A necessary token budget}

A substitution can change its own event and the successor's context, as
established in KGW and editing-robustness analysis
\citep{kirchenbauer2023watermark,zhao2024robust}. Accounting for global
occurrence cancellation gives the following length-sensitive bound.

\begin{proposition}[Support and historical margin capacity]
\label{prop:math-capacity}
For complete token paths at global token Levenshtein distance $q$,
\begin{equation}
D\le 2q-\ell_+,\qquad C\le 2q-(-\ell)_+,\qquad
D+C\le 4q-|\ell|.
\label{eq:math-support-bound}
\end{equation}
For fixed binary event states, $\gamma=1/2$, $\tau=4$, $T\ge4$ and $T'>0$,
\begin{equation}
\Delta M\le2q.
\label{eq:math-twoq}
\end{equation}
An initially positive source therefore requires
$q\ge\lfloor M_0/2\rfloor+1$ for strict escape.
\end{proposition}

The proof and general-$\gamma$ ceiling appear in
Appendices~\ref{app:math-proofs} and \ref{app:crossing-scales}.
This necessary bound grants every changed event its most favorable state.
It guarantees neither a faithful candidate nor favorable realized signs.
For an initially detected source in this configuration, $T>16$, so the
one-sided length condition holds automatically. A two-sided version also
certifies robustness to subsequent token edits within its stated domain
(Appendix~\ref{app:math-stability}).

\subsection{Source enrichment and estimated starting strength}
\label{sec:starting-strength}

Watermarked sampling enriches the source in green events. Prior color-agnostic
editing analysis already distinguishes enriched retained fragments from nominal
new fragments \citep{wu2024color}. Under the additional
approximation that destroyed events have common positive centered mean
$\epsilon$ and created evidence has conditional mean zero,
\begin{equation}
\mathbb E[L\mid\mathcal I_{\mathrm{edit}},x']\approx\epsilon D.
\label{eq:math-reset-reference}
\end{equation}
Here $\mathcal I_{\mathrm{edit}}$ is the editor's permitted information, including its chosen
candidate. This conditional reference motivates coverage selection; it supplies
neither a universal bound nor calibrated prediction. Created identities can
reuse observed contexts, and the active modulo-reseed implementation does not
justify secure-PRF independence.

Initial-strength estimation addresses the required reduction in
Eq.~\eqref{eq:math-crossing}. The historical diagnostic in
Section~\ref{sec:historical-information} asks whether richer observations reveal
this headroom. Let
$\mathcal I_{\mathrm{trace}}$ include generation-time logits and known sampling
processors, which are unavailable to the final constructor.
Write $Z_0=Z_K(x)$. If
$m_e=\mathbb E[y_K(e)\mid\mathcal I_{\mathrm{trace}}]$, linearity gives
\begin{equation}
\overline Z_0=\mathbb E[Z_0\mid\mathcal I_{\mathrm{trace}}]
=\frac{\sum_e n_x(e)m_e}{\sigma\sqrt T},\qquad
\widehat Z_0=\frac{\sum_e n_x(e)\widehat m_e}{\sigma\sqrt T}.
\label{eq:starting-score-estimate}
\end{equation}
The implemented $\widehat m$ approximates these conditional means.
This trace-assisted setting uses no key but has a richer observation contract.

At unchanged length with homogeneous enrichment, nominal created evidence,
$q,D>0$ and approximately constant efficiency $\eta=D/(2q)\le1$, the mean-response
reference becomes
\begin{align}
\mathbb E[\Delta Z\mid\mathcal I_{\mathrm{trace}},x']&\approx
\frac{D}{T}\widehat Z_0,\nonumber\\
q_{\mathrm{mean}}&\approx\frac{T}{2\eta}
\left(1-\frac{\tau}{\widehat Z_0}\right),\qquad\widehat Z_0>\tau.
\label{eq:strength-budget-reference}
\end{align}
This connects aggregate strength to the amount of evidence that editing must
replace. Candidate-specific source yield, created states and semantic
constraints can change the realized requirement. Moreover, estimating the sum
along the source and comparing two edits require different posterior directions:
shared evidence cancels in candidate contrasts. At equal output length and fixed
candidate-effect estimates, the initial-score term is common to all candidates
and cannot change their ranking. It can inform the separate question of whether
the predicted reduction is sufficient. The construction below does not use
$\widehat Z_0$ as a stopping rule. We evaluate the historical estimate and a
separate stopping attempt under their respective access contracts. Joint search
first constructs the attainable coverage choices at each actual cost; recognizing
a sufficient choice additionally requires information about its signed effect.

%% file: figures/figure_one_path.tex
\begingroup
\input{figures/figure_one_styles}
\begin{tikzpicture}[fonebase]
\path[use as bounding box] (0,0.45) rectangle (13.7,2.57);
\node[anchor=west] at (0.08,1.00) {Source $x$};
\node[anchor=west] at (0.08,2.00) {Output $x'$};
\node[fonetoken] (sa) at (2.30,1.00) {$a$};
\node[foneoriginal] (sb) at (4.10,1.00) {$b$};
\node[foneoriginal] (sc) at (5.90,1.00) {$c$};
\node[fonetoken] (sd) at (7.70,1.00) {$d$};
\node[fonetoken] (oa) at (2.30,2.00) {$a$};
\node[fonerewritten] (ou) at (4.10,2.00) {$u$};
\node[fonerewritten] (ov) at (5.90,2.00) {$v$};
\node[fonetoken] (od) at (7.70,2.00) {$d$};
\foreach \a/\b in {sa/sb,sb/sc,sc/sd}{
  \draw[fonestateG] (\a.east)--node[above=3pt,text=foneGreen]{G}(\b.west);
}
\draw[fonestateG] (oa.east)--node[below=3pt,text=foneGreen]{G}(ou.west);
\draw[fonestateR] (ou.east)--node[below=3pt,text=foneRed]{R}(ov.west);
\draw[fonestateR] (ov.east)--node[below=3pt,text=foneRed]{R}(od.west);
\foreach \name/\y in {sa/1.00,oa/2.00}{
  \node at (1.56,\y) {$\cdots$};
  \draw[draw=foneGray!60,dashed] (1.78,\y)--(\name.west);
}
\foreach \name/\y in {sd/1.00,od/2.00}{
  \draw[draw=foneGray!60,dashed] (\name.east)--(8.23,\y);
  \node at (8.45,\y) {$\cdots$};
}
\node[font=\fontsize{9}{10}\selectfont\itshape] at (5.00,1.51)
  {ordinary rewording; unchanged context};
\node[anchor=west] at (9.12,0.78) {Schematic states, $\gamma=1/2$};
\node[anchor=west,text=foneGreen] at (9.12,1.12) {G: $+1/2$};
\node[anchor=west,text=foneRed] at (11.38,1.12) {R: $-1/2$};
\node[anchor=west] at (9.12,1.62) {Editor sees token paths.};
\node[anchor=west] at (9.12,1.98) {Evaluator supplies colors.};
\node[anchor=west,text=foneGray] at (9.12,2.34) {Other events cancel.};
\end{tikzpicture}
\endgroup

%% file: figures/figure_one_styles.tex
\definecolor{foneSource}{HTML}{0072B2}
\definecolor{foneOutput}{HTML}{D55E00}
\definecolor{foneGreen}{HTML}{237A46}
\definecolor{foneRed}{HTML}{AD3035}
\definecolor{foneInk}{HTML}{333333}
\definecolor{foneGray}{HTML}{777777}
\tikzset{
  fonebase/.style={x=1cm,y=-1cm,>=Latex,
    font=\fontsize{9}{10.5}\selectfont,
    every node/.style={inner sep=0pt,text=foneInk}},
  fonetoken/.style={draw=foneGray!60,rounded corners=1.5pt,
    minimum width=0.64cm,minimum height=0.43cm,fill=white},
  foneoriginal/.style={fonetoken,draw=foneSource,fill=foneSource!6},
  fonerewritten/.style={fonetoken,draw=foneOutput,fill=foneOutput!6},
  fonestateG/.style={->,line width=1.0pt,draw=foneGreen},
  fonestateR/.style={->,line width=1.0pt,draw=foneRed}
}

%% file: sections/construction.tex
\section{Constructing and selecting local rewrites}
\label{sec:construction}

\subsection{Constructing the candidate language}
\label{sec:candidate-language}

Algorithm~\ref{alg:coverage-construction} constructs ordinary rewrites and
selects among their complete combinations under a token-edit budget. The
procedure first chooses phrase locations, requests alternatives and applies
explicit mechanical checks. It then measures complete outputs so that shared
token boundaries and interactions between replacements enter the selection.
A fixed rule chooses at most eight disjoint source-character spans using
punctuation, word-count limits and distributed quantiles. GPT-5.6 Luna at low
reasoning effort receives the whole source and two requests per span: a small
wording change and a grammatical reordering/equivalent phrase. Instructions
preserve propositions, roles, referents, quantities, qualifications, modality
and negation; the provider receives no tokenizer objective, key or score.

Mechanical admission checks each single-site complete output for canonical
round-trip, token distance between one and eight, absolute net length change
at most $0.05N$, lexical anchors and ordinary formatting. Malformed or missing
responses remain failures without automatic retries. These literal checks
precede global preservation review. Exact rules and the language $\mathcal L(x)$ appear in
Appendices~\ref{app:construction-details} and \ref{app:selection-comparators}.

Let $\operatorname{RT}(x')$ mean that decoding the canonical encoding of the
complete composed prose reproduces that prose exactly. For
$\rho\in\{0.05,0.10,0.15,0.20\}$, feasible compositions satisfy
\begin{equation}
  \mathcal{F}_{\rho}(x)=
  \left\{x'\in\mathcal{L}(x):
    q(x,x')\leq\lfloor\rho N\rfloor,\quad
    |\ell(x')|\leq 0.05N,\quad T'>0,\quad \operatorname{RT}(x')
  \right\}.
  \label{eq:construction-feasible}
\end{equation}

\subsection{Joint selection under actual token budgets}

To compare selectors at equal actual editing cost, we first identify the
largest attainable token-edit count within each allowance. Among outputs with
that count, coverage selection removes the most original event occurrences:
\begin{align}
  q_{\rho}&=\max_{x'\in\mathcal{F}_{\rho}(x)}q(x,x'),
  \label{eq:construction-dose}\\
  x'_{\rho}&=\underset{\substack{x'\in\mathcal{F}_{\rho}(x)\\
                     q(x,x')=q_{\rho}}}{\operatorname{lexmin}}
    \left(-D(x'),\ |\ell(x')|,\ s(x'),\ \mathbf{i}(x'),\ h(x')\right).
  \label{eq:construction-selector}
\end{align}

Here $s$ counts edited sites, $\mathbf i$ lists candidate IDs in source order,
and $h$ is the text hash; ties use the retained deduplication representative.
The largest-attainable-dose rule was frozen for matched-cost selector comparisons;
it supplies no minimum-cost crossing guarantee. Let $B[q]$ retain the winner of
the same coverage ordering at each attained actual cost. The returned output is
$B[q_\rho]$; this map separates joint selection from the dose decision.

\input{sections/algorithm_construction}

\subsection{Exact implementations and computational tradeoffs}

Equation~\eqref{eq:math-reset-reference} motivates this objective: removing
more watermark-enriched source events can reduce expected evidence under the
stated assumptions on their enrichment and the introduced events.
Exactness applies to coverage in the finite actual-cost
slice; semantic preservation and crossing remain measured outcomes.

Complete-output enumeration measures shared edges, boundary retokenization,
recreated identities and global alignment directly. A follow-up prefix-sharing
dynamic program (DP) computes the same $B[q]$ on the materialized token paths.
Its state retains an exact minimum-distance column and capped source-event
overlap; bottom-up tables choose the best descendant at each actual cost.
It reproduced all 420 allowance selections and 1,521 attained-cost choices
(Appendix~\ref{app:coverage-dp}). This verifies the coverage solver on the supplied
bank; every terminal is examined and upstream enumeration remains necessary
for this implementation. Graph construction and storage make its computational
benefit conditional (Appendix~\ref{app:solver-efficiency}). The historical
additive-utility DP has a different objective and is retained in the
technical companion.

There are at most $3^8=6561$ configurations, measured using a tokenizer and
bit-parallel edit distance, with no local language model or partition computation.
An isolated replay using source prose and recorded proposal responses reproduced
all 420 coverage outputs on 105 sources with networking and model/provider/detector
imports disabled. This verifies the local constructor's access boundary; fresh
proposal generation remains a separate provider operation.
Median local selection took $8.92$\,ms per source, excluding provider and review inference
(Appendix~\ref{app:reporting-controls}). The same-bank surprisal comparator
and its uncalibrated rank utility are specified in
Appendix~\ref{app:selection-comparators}.

%% file: sections/algorithm_construction.tex
\begin{algorithm}[H]
\caption{Budgeted token-event coverage construction}
\label{alg:coverage-construction}
\small
\begin{algorithmic}[1]
\Require Source prose with scorable, round-tripping token path $x$;
  pinned tokenizer; rewrite provider; frozen nomination and admission rules.
\Ensure Sealed outputs for $\rho\in\{0.05,0.10,0.15,0.20\}$ and structural metadata.
\State Count source adjacent-token occurrences $n_x$; set $N=|x|$.
\State Nominate, request and mechanically admit alternatives (Section~\ref{sec:candidate-language}).
\State Initialize seen-text hashes $H\gets\varnothing$ and cost-winner map $B\gets\varnothing$.
\For{each keep/alternative configuration in deterministic order}
  \State Compose complete prose $u$; set $h\gets\operatorname{SHA256}(u)$; \textbf{if} $h\in H$ \textbf{then continue}.
  \State Add $h$ to $H$; canonically encode $u$ as $x'$.
  \State Measure minimum token distance $q(x,x')$ and $\ell=|x'|-N$.
  \State \textbf{if} $x'\notin\mathcal F_{0.20}(x)$ \textbf{then continue}.
  \State Measure complete-output event-multiset differences $D,C$ using Eq.~\eqref{eq:math-incidence}.
  \State Retain $x'$ in $B[q]$ if its tuple in Eq.~\eqref{eq:construction-selector}
    is smaller, or $B[q]$ is empty.
\EndFor
\State Read all four $q_\rho$ from the completed map using Eq.~\eqref{eq:construction-dose}.
\State \Return sealed $B[q_\rho]$ and metadata for each budget; the identity is available at $q=0$.
\end{algorithmic}
\vspace{2pt}
\small Sites follow source order; each site lists keep first, then
alternatives by candidate ID. Text deduplication retains the first path.
Keyed scoring and blinded quality review occur after selection, without feedback.
\end{algorithm}

%% file: sections/experimental_design.tex
\section{Experimental design}
\label{sec:design}

\paragraph{Cohort and detector.}
We generated 128 watermarked continuations from 16 prompt groups crossed with
eight fresh keys. We edited these generated continuations. The saved prompts
came from a locally processed artifact described as derived from the Colossal
Clean Crawled Corpus (C4) \citep{raffel2020exploring}; its raw upstream split and URL lineage
remain unresolved. Qwen2.5-1.5B-Instruct \citep{qwen2025technical} used float32 CUDA, temperature $0.8$,
top-$k=20$, top-$p=0.95$, repetition penalty $1.1$, a four-gram repetition
constraint and a NumPy-PCG64 draw. Generation and detection shared the original
left-window KGW with $h=1$, $\gamma=0.5$, logit bias $\delta=2$ and terminal
threshold four. Scoring preserved original CUDA partitions.
Appendix~\ref{sec:appendix-evidence} records pinned revisions and hashes;
Appendix~\ref{app:execution} gives executed prompts and provider metadata.

Blinded source review preceded rewriting and keyed outcome evaluation.
Enrollment required a completed paragraph, at least 40 canonical tokens and
inherited language/coherence defect $B\le1$. It retained 105 sources, including
82 initially above $Z=4$, with median qualified length $N=85$. Sixteen
same-model unwatermarked passages supplied smoke controls. Every selector
used the same source-specific attained cost at each allowance; detector scores
did not determine dose.

\paragraph{Quality and outcomes.}
The frozen rubric separates benign variation $A$, inherited source defects $B$,
introduced factual/logical risk $C_{\mathrm{risk}}$, and introduced fluency
damage $F$. Relative quality requires $C_{\mathrm{risk}}=0,F\le1$;
polished quality further requires $F=0$, and absolute quality requires
$C_{\mathrm{risk}}=0,F=0,B=0$.
For confirmation, GPT-6.1 Sol with medium reasoning reviewed prose blinded to
scores and states.
Reuse required exact source/output hashes; missing judgments remained
unreviewed. These are model-only source-preservation judgments, without human
or independent factual validation.

The prospective primary compared weighted and coverage selection in
relative-quality strict escape at the $20\%$ allowance. Secondary controls
used deterministic hash ordering and weighting from a public 3B checkpoint.
The primary weighted rule used the public generating checkpoint.
Neither probability comparator received generation traces or keyed states.
Continuous original summaries use all 105 sources; cost--outcome curves use
the same 82 positives for both coordinates. Weighted outputs have complete
reviews at all four doses. Later completion of 25 missing coverage crossing
reviews determines that policy's qualified-escape curve, while noncrossing
outputs without judgments remain unreviewed (Appendix~\ref{app:dose-metrics}).

\paragraph{Inference and follow-ups.}
We retained within-document pairing and 10,000 crossed prompt/key bootstrap
replicates, seed 130913, over 16 prompt/eight key clusters, using independent
axis resampling as in the pigeonhole bootstrap \citep{owen2007pigeonhole}. Intervals are
pointwise; Appendix~\ref{app:bootstrap} specifies resampling and empty-draw
handling. Candidates, doses and repeated events do not increase the source
sample size. Retrospective completion of 39 hash labels preserved 66 existing
judgments and did not change the primary.

Subsequent audits enumerated the frozen bank, reviewed minimum-cost witnesses
and tested blind stopping with training/calibration excluding held-out prompt
and key groups. A solver check compared coverage DP with enumeration; a separate
reconstructed-observation analysis tested source-evidence selection at matched
actual cost. These analyses reuse previously analyzed sources. Historical strength
estimation, nomination and the ordinary-rewriting development comparison use
separate cohorts and access contracts
(Table~\ref{tab:cohort-roles}); none enlarges the confirmation sample.
Detailed protocols and missingness are retained in
Appendices~\ref{app:reporting-controls}, \ref{app:historical-information}
and \ref{app:stopping}.

%% file: sections/results.tex
\section{Results}
\label{sec:results}

\subsection{Budgeted construction produces reviewed terminal escapes}

We first test whether composing ordinary local rewrites can cross the terminal
threshold while meeting the token budget and source-preservation rubric.
At the $20\%$ allowance, coverage selection achieved 39/82
relative-quality-qualified escapes ($47.6\%$, crossed 95\% interval
$[28.6\%,66.7\%]$), at mean actual cost $15.94$ tokens ($15.96\%$)
among initially positive qualified sources. Of its 40 raw crossings,
39 passed relative review and 34 passed polished review.
Thus the construction yielded reviewed escapes in this setting, while the
stricter fluency criterion reduced their number. The small raw-to-relative gap
describes admitted outputs under model review; it does not measure human
fidelity or the effect of quality constraints on candidate supply.

Figure~\ref{fig:dose} distinguishes continuous suppression from threshold
crossing across editing allowances. Mean $\Delta Z$ rose from $0.436$ to
$1.373$, but a reduction produces escape only when it exceeds the source's
initial score excess. Most of the observed increase in qualified crossings
occurred by the $15\%$ allowance: the count rose from 13 at $5\%$ to 37 at
$15\%$, then to 39 at $20\%$. The final spending increase therefore added
few crossings in aggregate, without establishing noninferiority of the smaller
allowance. These are nonnested fixed-dose selections, so the curve supplies
neither individual monotonicity nor an optimal cost frontier.
Appendix~\ref{app:dose-metrics} retains all operating points, uncertainty,
the separate weighted curve and retrospective completion of 25 coverage reviews.

\input{figures/dose_panels}

Probability weighting did not improve the prospective primary endpoint.
Weighted and coverage selection each achieved 39/82 qualified escapes at
identical actual costs; their paired difference was zero percentage points
with 95\% interval $[-6.67,+6.45]$. This supports the feasibility of the
tokenizer-only selection stage within the shared construction, without
establishing equivalence or an advantage over ordinary constrained rewriting.
We examine those attribution limits below.

\subsection{Evidence replacement and initial margin account for the outcomes}

To identify what produces suppression, we separate removed source evidence,
created evidence and the length boundary. This mechanism analysis retains
frozen weighted-policy outputs across all 105 qualified sources, separately
from the coverage dose curve. At the largest dose, removed source evidence
contributed about $6.14$ margin units on average, compared with $0.20$ from
created states and a small adverse length contribution. The complete
decomposition is retained in Eq.~\eqref{eq:observed-margin-decomposition}.
Destroyed occurrences were $78.92\%$ green, whereas created occurrences were
$49.02\%$ green; mean absolute length change was only 1.44 tokens ($1.64\%$).
Replacing enriched source evidence therefore accounted for most of the mean
reduction. These evaluator-side summaries contain correlated events and do
not establish a blind predictor of individual signed effects.

Individual crossings also depend on which evidence is removed and introduced.
Figure~\ref{fig:mechanism-examples} tracks the remaining margin in two
retrospectively chosen outputs with similar initial margins. The successful
rewrite destroyed fewer events, yet removed more source evidence; its created
events further reduced the margin. In the still-detected rewrite, created
evidence offset half of the source contribution before the length penalty.
Their net centered-evidence removal was 10 versus 3. The first output ended
only slightly below the threshold, whereas the second retained a substantial
positive margin. Both passed relative and polished model review. Their
different sources and costs make this an accounting illustration, without a
matched causal or per-edit attribution claim.

\input{figures/mechanism_examples}

Initial headroom determines how much reduction is required. Among weighted
outputs, qualified escape counts were 28/37 for $4<Z_0\le5$, 10/28 for
$5<Z_0\le6$, and 1/17 for $6<Z_0\le8$. These descriptive strata place
38/39 successes at $Z_0\le6$, connecting the observed yield to source strength.
Neither this association nor the favorable-sign budget bound guarantees a
crossing for an individual source. Appendix~\ref{app:witness-diagnostics}
reports the full event, budget and post-edit stability diagnostics, including
the opening example's lack of a positive certified repair radius.

\subsection{The candidate bank separates selection limits from dose opportunities}
\label{sec:ordinary-baseline}

The next question is whether better choices within the admitted bank can
improve these outcomes. At fixed actual cost, weighting added mean
$\Delta Z=0.0196$ $[-0.0380,0.0883]$ over coverage; retrospective constant
and uniform-choice controls also left added selection value unestablished.
Choice was restricted: among 82 positive sources at the largest dose, only
six attained-cost menus contained both crossing and noncrossing outputs.
The null comparisons neither identify the effects of shared proposal and
admission stages nor establish selector equivalence. Full policy outcomes,
turnover and control intervals appear in Appendix~\ref{app:selection-controls}.

A separate comparison tests the broader construction against ordinary
localized rewriting. On 16 previously analyzed development sources, the
historical weighted constructive reference added mean $\Delta Z=0.011$
$[-0.486,0.509]$ at identical actual token cost; both arms escaped on 7/10
initially detected sources. All selected outputs passed relative review,
with mixed cached Sol/new Astra reviewers under a common rubric. Unequal
proposal supply and inference cost, source reuse and the wide interval leave
both superiority and equivalence unresolved. Appendix~\ref{app:ordinary-comparison}
retains the competence gate and the separate null all-cost-objective test.

Complete enumeration shows where a different dose could matter even when
fixed-cost selection has limited opportunity. The actual-key audit scored
35,548 eligible unique outputs, reproduced 1,680 original endpoints and
partitioned the 82 positives in Table~\ref{tab:bank-failure-partition}.
Of the 42 coverage failures, two could cross through another output at the
same actual cost, fourteen had crossings only at a lower cost, and twenty-six
had no crossing in the allowed bank. Improving ranking only at the largest
attained cost cannot recover the fourteen lower-cost-only opportunities.
The last category identifies a limit of this frozen candidate language for
the realized key, without implying universal editing impossibility.

\begin{table}[t]
\centering
\caption{Mutually exclusive outcomes for the 82 initially detected qualified
sources at the $20\%$ allowance. Counts use raw strict $Z'<4$, the actual key
and the frozen candidate language. The first row includes 39 reviewed
quality-qualified escapes; no quality qualification is imputed to oracle
alternatives at the same actual cost. Subsequent review qualified a minimum-cost
witness for each of the 56 bank-crossing sources, as reported below.
The source sets were checked directly against saved records.}
\label{tab:bank-failure-partition}
\begin{tabularx}{\linewidth}{Xr}
\toprule
Source outcome & Count \\
\midrule
Escaped under the original coverage selection & 40 \\
Additional escape available at the same actual cost & 2 \\
Additional escape available only at a lower actual cost & 14 \\
No escaping output in the full allowed candidate bank & 26 \\
\midrule
Total initially detected qualified sources & 82 \\
\bottomrule
\end{tabularx}
\end{table}

Quality review confirms that cheaper crossings can preserve the source under
the rubric. All 56 passed relative quality and 51 were polished. Their costs
therefore attain both raw and relative-quality
crossing minima within this bank. On the same 39 frozen coverage successes,
mean cost was $17.31\%$ for the policy and $5.43\%$ for reviewed minima.
This actual-key-discovered gap establishes unused cost-saving opportunities,
without changing the operational 39/82 result. Many witnesses crossed narrowly:
median slack was $4-Z'=0.120$, with 40/56 between $3.8$ and four.
These small slacks make prediction error consequential for a rule seeking
to identify the cheaper crossings without detector access. This motivates
the final analysis.

\subsection{Estimating headroom and recognizing sufficient editing}
\label{sec:recognition}

\paragraph{Information about initial strength.}
\label{sec:historical-information}

For the evidence-based stopping formulation studied here, remaining margin
combines initial headroom with a candidate's signed effect. Estimating initial headroom is one route
to reducing this uncertainty. On the separate historical 95-source CUDA
cohort, trace-assisted $\widehat Z_0$ correlated $0.793$ with measured
$Z_0$, with mean error $+0.089$ and error standard deviation $0.769$.
An in-sample regression on an intercept and $\sqrt T$ had $R^2=0.494$;
adding $\widehat Z_0$ raised it to $0.630$
(Figure~\ref{fig:starting-strength}). The generation observations therefore
contained aggregate strength information beyond length in this exposed
cohort, with no held-out calibration or plaintext prediction claim.

\input{figures/starting_strength_panels}

This association did not establish added allocation value: retrospective
length-first and estimate-first allocation each yielded 32 crossings,
compared with 33 for saved triage at slightly different expenditures
(Appendix~\ref{app:historical-information}). The final constructor uses no
generation traces. The following stopping experiment evaluates a different
estimator and cohort, so its outcome does not test the historical estimator.

\paragraph{From initial strength to candidate choice.}
A follow-up reconstructed generator observations for the confirmation bank to
test whether inferred source evidence improves selection. With the bank,
enumerator and actual edit costs fixed, maximizing predicted destroyed evidence
added mean $\Delta Z=0.0314$ with crossed 95\% interval $[-0.0372,0.0988]$
across 104 resolved sources; one source used coverage fallback. This richer
information did not establish an advantage over coverage. Two newly selected
crossings lacked quality reviews. Appendix~\ref{app:reconstructed-selection}
retains the operational results and quality bounds. Estimating aggregate
strength and predicting differences between rewrites therefore remain distinct
empirical tasks; exact search does not ensure an accurate signed-effect objective.

\paragraph{Recognizing a sufficient dose without traces.}
Returning to the 105-source confirmation bank, a blind path keeping one
coverage-selected output per attainable cost retained raw crossings for
50/56 bank-crossing sources, attaining the full-bank minimum cost for 31.
These counts establish substantial retained opportunity, without implying
identity with the reviewed witnesses. A controller estimated remaining
evidence from length and public-model mean surprisal, using held-out
prompt/key groups, other-key training scores and an empirical upper-residual
allowance. Its information differs from both the original constructor's
and the historical trace-assisted estimator's.

The primary uncertainty-adjusted rule selected
exactly the original $20\%$-allowance coverage outputs for all 105 sources.
Only one of the 82 positives met its predicted-target condition; 81 returned
the fixed-dose fallback. Unchanged yield thus reflects fallback, with only
1/82 positive-source target declarations. Removing the uncertainty allowance
saved $4.41$ cost percentage points ($15.96\%\to11.55\%$; 95\% interval
$[2.24,6.90]$), but raw crossings fell from 40 to 25. The cheaper outputs
exposed a cost--crossing tradeoff, without demonstrating preserved quality
and robustness (Appendix~\ref{app:stopping}).

Starting-score error and intervention-conditioned signed effects both
contributed to remaining-score error; neither dominated uniformly.
The narrow slack of many cheap witnesses leaves little tolerance for their
combined error. The tested controller did not convert retained candidate
opportunities into reliable savings. Reliable additional cost savings over
the fixed-dose rule remain unresolved.
The reviewed witnesses nevertheless establish that the largest-dose rule
does not minimize actual-key crossing cost within this bank. This separates
the demonstrated ability to construct budgeted escapes from the unresolved
ability to recognize the cheapest sufficient edit without detector access.

%% file: figures/dose_panels.tex
\begin{figure}[t]
  \centering
  \begin{subfigure}{0.49\linewidth}
    \includegraphics[width=\linewidth]{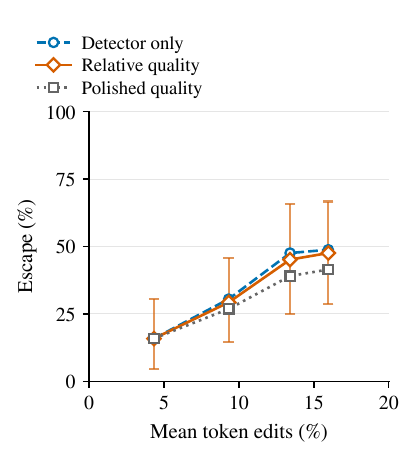}
    \caption{Strict crossing and quality.}
  \end{subfigure}\hfill
  \begin{subfigure}{0.49\linewidth}
    \includegraphics[width=\linewidth]{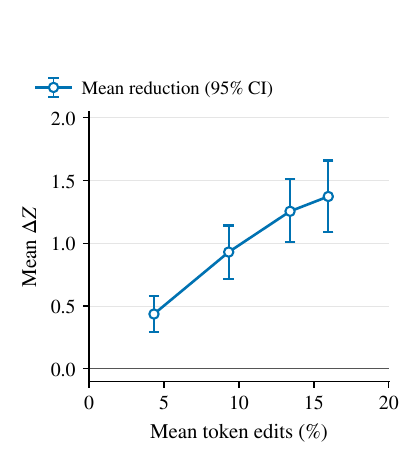}
    \caption{Continuous score reduction.}
  \end{subfigure}
  \caption{Frozen coverage-policy operating points under 5\%, 10\%, 15\% and
  20\% allowances, in increasing horizontal order. Both axes in both panels
  use the same 82 initially detected qualified sources. Actual cost is mean
  global token distance divided by source token count. Panel (a) separates
  detector-only, relative-quality and polished-quality escapes; panel (b)
  retains all score changes in the mean. Error bars are pointwise crossed
  prompt/key 95\% intervals for relative-quality escape and mean $\Delta Z$
  (16 prompt/eight key clusters). Outputs are nonnested fixed-dose selections,
  and no dose was chosen by detector score. Retrospective review completed
  25 previously unreviewed crossing outputs under the unchanged rubric;
  all crossing identities now have judgments.}
  \label{fig:dose}
\end{figure}

%% file: figures/mechanism_examples.tex
\begin{figure}[t]
  \centering
  \begin{subfigure}[t]{0.48\linewidth}
    \centering
    \input{figures/figure_one_escape}
    \caption{Successful illustration.}
    \label{fig:mechanism-escape}
  \end{subfigure}\hfill
  \begin{subfigure}[t]{0.48\linewidth}
    \centering
    \input{figures/figure_one_detected}
    \caption{Still-detected illustration.}
    \label{fig:mechanism-detected}
  \end{subfigure}
  \caption{Remaining detector margin for two retrospectively chosen weighted
  selections at the $20\%$ allowance. Both panels use the same margin scale;
  strict terminal escape requires a final value below the dashed zero line.
  Starting at $M_0$, source removal changes the margin by $-A_D$ (blue),
  created evidence by $+A_C$ (orange), and length by $-\Delta b$ (gray),
  giving $M'=M_0-A_D+A_C-\Delta b$ (black). Numbers above intermediate bars
  are signed changes; initial and final bars show levels. These are accounting
  stages of complete outputs, not sequential edit scores or separate per-edit
  effects. Both outputs passed relative and polished model review. The
  successful output is the passage in Figure~\ref{fig:examples}(a).
  The different sources and costs preclude a matched causal comparison.}
  \label{fig:mechanism-examples}
\end{figure}
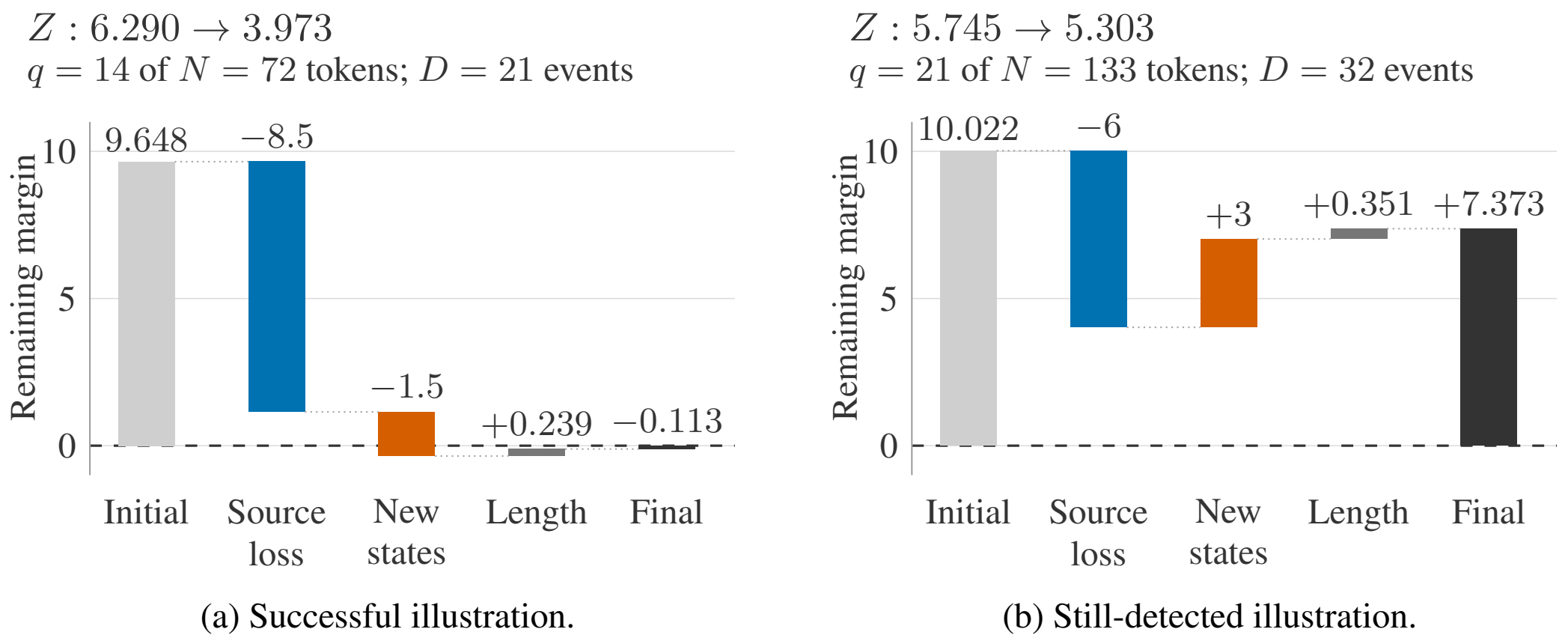

%% file: figures/figure_one_escape.tex
\begingroup
\def\FigOneRole{Escape}
\def\FigOneEscapeRole{Escape}
\input{figures/figure_one_outcome}
\endgroup

%% file: figures/figure_one_detected.tex
\begingroup
\def\FigOneRole{Detected}
\def\FigOneEscapeRole{Escape}
\input{figures/figure_one_outcome}
\endgroup

%% file: figures/starting_strength_panels.tex
\begin{figure}[t]
  \centering
  \begin{subfigure}{0.49\linewidth}
    \includegraphics[width=\linewidth]{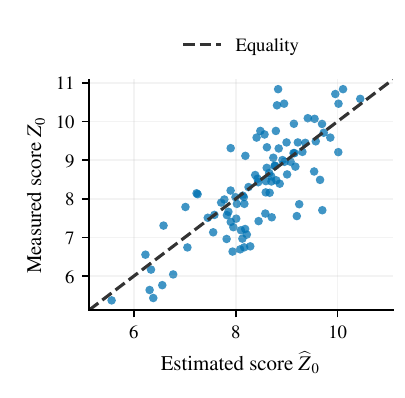}
    \caption{Aggregate strength, correlation $0.793$.}
  \end{subfigure}\hfill
  \begin{subfigure}{0.49\linewidth}
    \includegraphics[width=\linewidth]{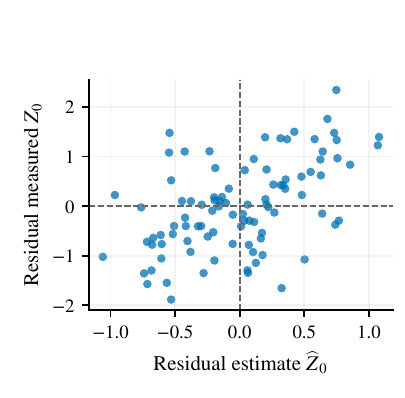}
    \caption{Length-adjusted association, $0.519$.}
  \end{subfigure}
  \caption{Historical trace-assisted starting-strength diagnostic on 95 sources
  (12 prompt/eight key clusters), separate from the closing confirmation.
  Panel (a) compares approximate key-free estimates with saved actual-key scores;
  the dashed line is equality. Panel (b) residualizes both quantities on an
  intercept and $\sqrt T$ in the same cohort; dashed axes indicate zero.
  Each point is one source. The association is descriptive, without held-out
  calibration or demonstrated allocation benefit beyond length. Stored
  generation observations are inputs to this diagnostic only, and are absent
  from the final constructor.}
  \label{fig:starting-strength}
\end{figure}

%% file: sections/discussion.tex
\section{Discussion and limitations}
\label{sec:discussion}

Complete-output measurement makes the editing constraint verifiable; the
weighted-policy analysis connects suppression to removed and created evidence.
The bank audit separates cheaper candidates' availability from their recognition:
the tested stopping failure supplies no cost lower bound for every key-blind method.
Coverage DP shares computation while preserving selections. Improving signed
prediction or candidate supply requires separate evidence.

The constructor can operate after target-model access ends, given the source,
known tokenizer and independent rewrite proposals. Observability through other
records or subsequent examination of edited text was not evaluated.

Further evaluation should distinguish the contributions of proposal generation,
admission, selection and dose. The present controls leave those shared stages
unseparated and competitive advantage unresolved. Initial-strength estimation
and candidate comparison require different information. The development
$Z_0$ result supports aggregate strength inference under its trace contract;
candidate-difference prediction, plaintext-only inference and added allocation
value beyond length require separate evidence. These distinctions constrain
which improvements the existing results can justify.

Quality and scope also delimit the result. The reviewer assesses preservation
of the source's claims; it can accept an output that retains an unsupported
source assertion. The exploratory Dolly pilot had no sources supported by the
supplied context, leaving the escape rate conditional on such sources
undefined (Appendix~\ref{app:additional-outcomes}). The confirmation covers short
sources and the original occurrence-counted $h=1$ KGW setting. Other generators,
context widths, seeding rules, repeat handling and prefix/interval detectors
require separate evaluation. Sixteen prompt and eight key clusters limit
precision, and 16 unwatermarked smoke controls cannot validate a rare-tail
false-positive rate. Local timing excludes proposal inference and quality
review. Fresh, component-matched comparisons and blinded human review would
address these remaining efficacy and fidelity boundaries.

%% file: sections/conclusion.tex
\section{Conclusion}
\label{sec:conclusion}

Jointly composing ordinary local rewrites produced
39/82 terminal KGW escapes that passed
source-relative model review, at mean actual token cost $15.96\%$ under a
$20\%$ allowance. Evidence accounting relates these outcomes to removed
source states, created states and initial headroom. Enumeration and a verified
prefix-sharing DP implement the same finite coverage rule. Controls established
no advantage for elaborate selection or over ordinary constrained rewriting.
Reviewed cheaper crossings demonstrate further candidate-bank feasibility.
Development $Z_0$ estimation provides partial headroom information from generation
observations; the separate no-trace stopping test leaves sufficient-dose
recognition unresolved. Actual cost, continuous reduction, crossing and
preservation together delimit the construction's demonstrated value.

%% file: sections/appendix_proofs.tex
\section{Evidence accounting, capacity and joint-composition proofs}
\label{app:math-proofs}

\subsection{Occurrence identities and exact score changes}

The event histogram $n_x(e)$ counts occurrences of the fixed ordered identity
$e$. Define $a_e=n_x(e)-n_{x'}(e)$, $d_e=[a_e]_+$ and $c_e=[-a_e]_+$. Since
$a_e=d_e-c_e$, summing over identities gives
\begin{equation}
\sum_e(c_e-d_e)=T'-T=\ell,
\quad
\sum_e a_e y_K(e)=A_K(x)-A_K(x')=A_D-A_C.
\label{eq:math-app-cancel}
\end{equation}
The equality holds with repeated identities and with identities moved between
positions. It relies on occurrence-counting with fixed state per identity; a
repeat-deduplicated detector has a different histogram and statistic. For
$T,T'>0$, subtracting the two centered margins directly proves
Eq.~\eqref{eq:math-margin-identity}. For the normalized score, write
\begin{align}
Z_K(x)-Z_K(x')
 &=\frac{A_K(x)}{\sigma\sqrt T}
   -\frac{A_K(x)-L}{\sigma\sqrt{T'}}\nonumber\\
 &=\frac{L}{\sigma\sqrt{T'}}
   +Z_K(x)\left(1-\sqrt{T/T'}\right),
\label{eq:math-app-score}
\end{align}
which proves Eq.~\eqref{eq:math-score-identity}. The strict crossing criterion
follows by writing $M_K(x')=M_0-\Delta M$ and preserving the strict inequality.

\subsection{Proof of the support and capacity proposition}

For scorable endpoints, fixed binary event states, $0<\gamma<1$ and $\tau\ge0$,
\begin{align}
L&\le(1-\gamma)D+\gamma C,\label{eq:math-evidence-ceiling}\\
\Delta M&\le 2q-(1-\gamma)\ell_+-\gamma(-\ell)_+
 +\tau\sigma\bigl(\sqrt{T+\ell}-\sqrt T\bigr).
\label{eq:math-margin-ceiling}
\end{align}
The following proof establishes this general ceiling and the historical
corollary stated in the main text.

\begin{proof}[Proof of Proposition~\ref{prop:math-capacity}]
For nonnegative counts $u,v,w$, $[u-w]_+\le[u-v]_++[v-w]_+$. Coordinatewise
application proves a directed multiset triangle inequality
$D(x,z)\le D(x,y)+D(y,z)$, and the same inequality holds for $C$ with the
arguments reversed. It remains valid when paths contain repeated identities or
an intermediate path is empty.

An internal substitution destroys at most two adjacent events and creates at
most two. A deletion destroys at most two and creates at most one; an insertion
destroys at most one and creates at most two. Endpoint operations and identity
cancellation can only decrease these counts. Consider a shortest token edit
script with $s$ substitutions, $r_-$ deletions and $r_+$ insertions, where
$q=s+r_-+r_+$. The directed triangle inequalities along this script give
\begin{align}
D&\le2s+2r_-+r_+=2q-r_+,\nonumber\\
C&\le2s+r_-+2r_+=2q-r_-.
\label{eq:math-app-script}
\end{align}
Since $\ell=r_+-r_-$, we have $r_+\ge\ell_+$,
$r_-\ge(-\ell)_+$ and $r_++r_-\ge|\ell|$, proving
Eq.~\eqref{eq:math-support-bound}. The support inequalities also hold for empty
endpoints if event count is defined as $(N-1)_+$. The identity $C-D=\ell$
uses nonempty endpoints, as in the scorable domain of the score equations.

Every destroyed centered state is at most $1-\gamma$, and the negative of every
created centered state is at most $\gamma$. Hence
$L\le(1-\gamma)D+\gamma C$. Applying the first two support bounds to this
weighted sum, followed by Eq.~\eqref{eq:math-margin-identity}, gives
Eq.~\eqref{eq:math-margin-ceiling}. This is a relaxation: states of shared or
constrained partitions need not simultaneously attain all favorable extremes.

At $\gamma=1/2$ and $\tau=4$, the right side simplifies to
\begin{equation}
2q-\frac{|\ell|}{2}+2\bigl(\sqrt{T+\ell}-\sqrt T\bigr).
\label{eq:math-app-historical-bound}
\end{equation}
If $\ell\ge0$, square-root concavity and $T\ge4$ imply
$2(\sqrt{T+\ell}-\sqrt T)\le\ell/\sqrt T\le\ell/2$. If $\ell<0$, both
the absolute-length penalty and square-root change are negative. Thus
$\Delta M\le2q$. Strict escape requires $\Delta M>M_0$ and therefore
$2q>M_0$, yielding the stated integer floor.
\end{proof}

The support factor two is sharp in token space when the vocabulary contains
at least $3q+1$ distinct IDs. Start with $2q+1$ distinct
tokens and replace each even-positioned token by a distinct fresh ID. The
changed positions are internal and pairwise nonadjacent, their $2q$ original
incident edges disappear, and all new identities are fresh. Consequently
$D=C=2q$. The proved bound also establishes that the endpoints have minimum
distance at least $q$, while the displayed substitutions supply a script of
cost $q$. This witness establishes geometric sharpness without asserting a
semantic or fluency property.

For general $\gamma$ and $\tau\ge0$, the same $\Delta M\le2q$ corollary follows
whenever
\begin{equation}
T\ge\frac{\tau^2\gamma}{4(1-\gamma)}.
\label{eq:math-app-general-domain}
\end{equation}
Indeed, the positive-length boundary term is at most
$\tau\sigma\ell/(2\sqrt T)$, which under this condition is at most
$(1-\gamma)\ell$. For negative $\ell$ the boundary term is already adverse.
This sufficient condition makes explicit why a uniform margin constant depends
on the threshold, green fraction and scored-length domain. Context-width,
seeding-by-position and repeat-handling changes require corresponding analyses.

For a strictly detected source, $M_0>0$ and $A_0\le(1-\gamma)T$ imply
$T>\tau^2\gamma/(1-\gamma)$ for $0<\gamma<1$ and $\tau\ge0$.
The domain in Eq.~\eqref{eq:math-app-general-domain} is therefore automatic for
such sources; in the historical configuration, detection requires $T>16$.
The one-sided $\Delta M\le2q$ ceiling consequently applies to every strictly
detected source in this fixed-state adjacent-event setting, with a scorable
output. The length assumptions below remain explicit for the two-sided repair
statement because an escaped path can have negative margin.

\subsection{A two-sided margin and strict repair certificate}
\label{app:math-stability}

\begin{corollary}[Sufficient post-edit token stability]
\label{cor:math-stability}
For the historical $\gamma=1/2$, $\tau=4$ detector with fixed adjacent-token
states, two paths $u,v$ with at least four scored events each satisfy
\begin{equation}
|M_K(u)-M_K(v)|\le2d_{\mathrm{Lev}}(u,v).
\label{eq:math-app-lipschitz}
\end{equation}
If $v$ is an escaped output of margin $M'<0$ and scored length $T'\ge4$, every
further canonical-token edit of cost at most
\begin{equation}
r_{\mathrm{safe}}=
\min\left\{\left\lceil\frac{-M'}{2}\right\rceil-1,\ T'-4\right\}
\label{eq:math-app-safe-radius}
\end{equation}
retains strict escape.
\end{corollary}

\begin{proof}
Apply Eq.~\eqref{eq:math-twoq} to both ordered pairs $(u,v)$ and $(v,u)$.
For a repair to $v$ of distance at most $r$, the resulting margin is at most
$M'+2r$. The first radius constraint ensures this expression is strictly
negative. The second retains at least four scored events, since a distance-$r$
edit changes token count by at most $r$.
\end{proof}

The ceiling-minus-one preserves strictness. For example, $M'=-4$ certifies a
one-token repair; two edits can reach equality and are uncertified. In the
historical setting the strict condition can be checked without floating-point
threshold comparisons. For a proposed radius $r$, let $h=2G'-T'+4r$. Then
$M'+2r<0$ is equivalent to $h<0$ or $h^2<16T'$ when $h\ge0$. This is a
sufficient, evaluator-side decision certificate. Character- or word-level repairs
need their own canonical token distance, and the certificate supplies no
guarantee of semantic preservation.

\subsection{Conditional uncertainty and the crossing objective}
\label{app:math-uncertainty}

Conditional on editor information $\mathcal I$, let $y$ be the vector of centered
keyed states, $m=\mathbb E[y\mid\mathcal I]$ and
$\Sigma=\operatorname{Cov}(y\mid\mathcal I)$. For a fixed candidate with
incidence vector $a=n_x-n_{x'}$, its gain has
\begin{equation}
\mathbb E[\Delta M\mid\mathcal I]=a^\top m+\Delta b,
\qquad
\operatorname{Var}(\Delta M\mid\mathcal I)=a^\top\Sigma a.
\label{eq:math-app-gain-variance}
\end{equation}
The final margin instead has
\begin{equation}
\mathbb E[M'\mid\mathcal I]=n_{x'}^\top m-\tau\sigma\sqrt{T'},
\qquad
\operatorname{Var}(M'\mid\mathcal I)=n_{x'}^\top\Sigma n_{x'}.
\label{eq:math-app-final-variance}
\end{equation}
The unknown baseline and candidate gain are correlated because they involve
shared keyed states. Controlling gain variance consequently does not by itself
control the probability that $M'<0$. Likewise, two outputs with different
lengths can order differently under expected $Z$, expected margin and crossing
probability. The paper's measured construction optimizes its declared structural
objective; these conditional equations characterize the information gap without
asserting a calibrated chance-optimal policy.

Repeated identities also affect uncertainty directly. If distinct identities
were conditionally independent with nominal Bernoulli variance
$\gamma(1-\gamma)$, an event appearing $r_e$ times would contribute
$r_e^2\gamma(1-\gamma)$ to the variance of its occurrence-weighted sum.
Shared partition constraints and posterior observations can add covariance terms.
Candidate appearances and repeated occurrences therefore cannot be treated as
independent experimental units. The empirical intervals use the recorded
document, prompt and key structure.

\paragraph{Aggregate prediction and candidate contrasts.}
For candidates $i,j$ fixed by $\mathcal I$, write $n_i=n_{x'_i}$,
$a_i=n_x-n_i$ and $v_{ij}=a_i-a_j=n_j-n_i$. Their margin-gain contrast has
\begin{align}
\mathbb E[\Delta M_i-\Delta M_j\mid\mathcal I]
 &=v_{ij}^\top m+\Delta b_i-\Delta b_j,\nonumber\\
\operatorname{Var}(\Delta M_i-\Delta M_j\mid\mathcal I)
 &=v_{ij}^\top\Sigma v_{ij}.
\label{eq:math-app-contrast-variance}
\end{align}
For posterior-mean error $e_m=\widehat m-m$, aggregate starting-score error is
$n_x^\top e_m/(\sigma\sqrt T)$, whereas the error in the predicted margin-gain
contrast is $v_{ij}^\top e_m$. If the two directions are not collinear, zero
error in the former need not imply zero error in the latter. Shared states
cancel in the contrast, so absolute prediction error can also coexist with an
accurate candidate ordering. Under the homogeneous reset reference, its mean
is approximately $\epsilon(D_i-D_j)+\Delta b_i-\Delta b_j$; matching support
and length removes this common structural signal.

At equal output lengths, greater margin reduction and smaller final $Z$ induce
the same ordering. With different lengths, define
$w_{ij}=n_j/(\sigma\sqrt{T'_j})-n_i/(\sigma\sqrt{T'_i})$. Then
\begin{align}
\Delta Z_i-\Delta Z_j&=w_{ij}^\top y_K,\nonumber\\
\mathbb E[\Delta Z_i-\Delta Z_j\mid\mathcal I]&=w_{ij}^\top m,\nonumber\\
\operatorname{Var}(\Delta Z_i-\Delta Z_j\mid\mathcal I)
&=w_{ij}^\top\Sigma w_{ij}.
\label{eq:math-app-score-contrast}
\end{align}
These identities locate the information required for a selection decision.
They do not establish that the historical aggregate estimator or the final
surprisal utility estimates these directions reliably.

\paragraph{Position relative to prior editing bounds.}
The own-event/next-event locality and token-edit robustness arguments precede
this study \citep{kirchenbauer2023watermark,zhao2024robust}. Our exposition fixes
the scored count at $T=N-1$, carries occurrence multiplicity and net length
through the signed margin, and applies the resulting ceiling to the controlled
editing problem. Numerical constants for a normalized $Z$ cannot be compared
directly with an unnormalized centered score or a convention using all $N$
tokens in centering. We make no claim of the first editing-robustness theorem or
a generally tighter security bound.

\subsection{Equivalent crossing scales and conditional editing references}
\label{app:crossing-scales}

On the score scale, the crossing requirement in Eq.~\eqref{eq:math-crossing} is
\begin{equation}
Z'<\tau\quad\Longleftrightarrow\quad
\Delta Z>Z_0-\tau,\qquad
M_0=\sigma\sqrt T\,(Z_0-\tau),\quad Z_0=Z_K(x).
\label{eq:math-score-headroom}
\end{equation}
The reduction measures the intervention's effect, while $Z_0-\tau$ measures
the source's initial distance from crossing. This distinction motivates
reporting both continuous suppression and strict escape at a measured edit cost.
The observed score and its estimate have separate roles: keyed evaluation
supplies $Z_0$, and a blind estimate can only approximate the required headroom.

The required fraction is explicit when source evidence is positive. Writing
$A_0=A_K(x)>0$, $Z_0=Z_K(x)$ and $r_E=L/A_0$ gives
\begin{equation}
Z'=Z_0(1-r_E)\sqrt{T/T'},\qquad
Z'<\tau\ \Longleftrightarrow\
r_E>1-\frac{\tau}{Z_0}\sqrt{T'/T}.
\label{eq:math-evidence-fraction}
\end{equation}
Here $r_E$ is a realized signed evidence fraction and can lie outside $[0,1]$.
It differs from the structural fraction $D/T$: removed source states and created
states determine how much evidence a given support actually replaces. This
identity relates source strength to the required intervention without assuming
a homogeneous source or nominal created states.

\paragraph{A schematic equal-support contrast.}
Consider an internal substitution $(a,u,b)\to(a,v,b)$ with no identity collisions.
It has $q=1$, $D=C=2$ and $\ell=0$. At $\gamma=1/2$, destroying two green events
gives $A_D=1$. If the two created events are red, then $A_C=-1$ and $L=2$; if
both are green, then $A_C=1$ and $L=0$. This mathematical schematic isolates the
role of signed conversion at identical edit cost, support and length. The
measured examples in Section~\ref{sec:results} provide separate, retrospectively
selected illustrations with their recorded quality labels.

\paragraph{Structural efficiency and signed conversion.}
For
$q>0$ and $D=C>0$, let $\epsilon_D=A_D/D$ and $\epsilon_C=A_C/C$. Then
\begin{equation}
\Delta Z=\frac{q}{\sigma\sqrt T}\,
\frac{D}{q}\,(\epsilon_D-\epsilon_C).
\label{eq:math-efficiency}
\end{equation}
Structural efficiency satisfies $D/q\le2$, whereas the realized contrast
$\epsilon_D-\epsilon_C$ determines conversion of that support into score
reduction. The quantities on the right are evaluator diagnostics apart from
$q,D,T$. This decomposition explains how improved coverage can coexist with
weaker signed reduction and why additional structural search has diminishing
room when $D/q$ approaches two.

The mean-cost reference in Eq.~\eqref{eq:strength-budget-reference} assumes
approximately constant $\eta$ and nominal created evidence. It predicts a mean
response, with separate assumptions from the necessary favorable-sign ceiling.
All finite computations retain exact square roots. The technical companion
preserves the normalization expansions and their remainder bounds.

\subsection{Why complete compositions require joint measurement}
\label{sec:appendix-dp}

\input{figures/joint_edges}

In the distinct-identity schematic of Figure~\ref{fig:joint-edges}, use
$y_{ab}=y_K((a,b))$ and analogous subscripts. The isolated and composed gains are
\begin{align}
L_b&=y_{ab}+y_{bc}-y_{au}-y_{uc},\nonumber\\
L_c&=y_{bc}+y_{cd}-y_{bv}-y_{vd},\nonumber\\
L_{bc}&=y_{ab}+y_{bc}+y_{cd}-y_{au}-y_{uv}-y_{vd}.
\label{eq:construction-joint-gains}
\end{align}
Consequently,
\begin{equation}
L_{bc}-L_b-L_c=y_{uc}+y_{bv}-y_{bc}-y_{uv}.
\label{eq:construction-joint-correction}
\end{equation}
The correction removes a double-counted source edge and substitutes the actual
composed edge for the two isolated middle edges. It can have either sign.
The equal-length score correction divides this expression by $\sigma\sqrt T$.
This motivates whole-output measurement without establishing a solver advantage.

Three representation requirements also matter. Canonical tokenization must
be verified on complete prose because separately encoded fragments can merge
across boundaries. Actual cost is minimum Levenshtein distance; a nonminimal
alignment's cost cannot identify an exact-cost slice. Finally, coverage is
\begin{equation}
  D(x')=T-\sum_e\min\{n_x(e),n_{x'}(e)\}.
  \label{eq:construction-dp-coverage}
\end{equation}
It depends on capped occurrence overlap. Last-token state suffices for an
additive edge utility, but generic exact coverage also needs overlap history.
Enumeration measures all three quantities directly in the small final language.
The technical companion retains the alternative additive-utility DP, sufficient
state conditions and estimator-error bound. That formulation was not the
executed Algorithm~\ref{alg:coverage-construction}.

\subsection{Verified coverage DP on canonical output paths}
\label{app:coverage-dp}

The follow-up solver compiles the admitted canonical paths into a prefix trie.
At prefix $u$, it retains the exact Levenshtein column
$r_u(i)=\operatorname{Lev}(x_{1:i},u)$ for $0\le i\le N$, represented by
bit vectors, and capped counts $h_u(e)=\min\{n_x(e),n_u(e)\}$.
The prefix identifies its length and last token. Appending token $v$ uses the
ordinary minimum-distance recurrence and increments $h_u(e)$ only when the new
edge $e$ has not exhausted its source multiplicity. At a terminal, actual cost
is $r_u(N)$ and $D=T-\sum_e h_u(e)$.

Let $J(u,q)$ be the best descendant terminal at actual cost $q$, ordered by
the tuple in Eq.~\eqref{eq:construction-selector}. The bottom-up recurrence is
\begin{equation}
J(u,q)=\operatorname{lexmin}
\left(\mathcal T(u,q)\cup
\{J(uv,q):uv\text{ is a child of }u\}\right),
\label{eq:coverage-trie-dp}
\end{equation}
where $\mathcal T(u,q)$ contains admissible outputs ending at $u$ with actual
cost $q$; unavailable entries are omitted. The root table is $B[q]$.

Exactness follows by induction. The empty prefix initializes the distance
column and zero overlap. Each emission preserves the exact minimum distance
and capped event intersection, including repeated identities. Consequently,
terminal costs and coverages equal independent whole-output measurements.
Taking the minimum over each subtree preserves the best complete output at
every actual cost, including all deterministic ties. Reading $B[q_\rho]$
therefore reproduces the enumeration policy. Unlike an arbitrary alignment
cost, $r_u(N)$ identifies the required minimum-distance slice.

The implementation uses depth-first rollback for counts and visits every
terminal. It merges common prefixes, without suffix-state compression or
pruning based on predicted watermark states. Agreement on the supplied bank
does not establish an efficient compact representation of all possible prose
rewrites. Canonical paths are materialized before compilation; candidate
generation and its upstream enumeration remain outside this solver comparison.

%% file: figures/joint_edges.tex
\begin{figure}[t]
  \centering
  \begin{tabular}{lll}
    \toprule
    Schematic path & Token sequence & Adjacent-token identities \\
    \midrule
    Source & $a\ b\ c\ d$ & $(a,b),\ (b,c),\ (c,d)$ \\
    Replace $b$ & $a\ u\ c\ d$ & $(a,u),\ (u,c),\ (c,d)$ \\
    Replace $c$ & $a\ b\ v\ d$ & $(a,b),\ (b,v),\ (v,d)$ \\
    Replace both & $a\ u\ v\ d$ & $(a,u),\ (u,v),\ (v,d)$ \\
    \bottomrule
  \end{tabular}
  \caption{Schematic of a joint-composition interaction, with distinct token
  identities and unchanged tokenization. The isolated edits each have $q=1$
  and $D=2$; the joint path has $q=2$ and $D=3$. Adding isolated counts would
  remove the source edge $(b,c)$ twice and omit the new joint edge $(u,v)$.
  No colors, measured detector effect or semantic quality are assigned.
  Appendix~\ref{sec:appendix-dp} derives the signed correction.}
  \label{fig:joint-edges}
\end{figure}

%% file: sections/appendix_evidence.tex
\section{Construction, evaluation and reproducibility}
\label{sec:appendix-evidence}
\label{app:reporting-controls}

\subsection{Cohorts, configuration and information access}

The confirmation contains 128 attempted generations, 105 prospectively
qualified sources and 1,680 policy-dose records. Its locally processed
C4-derived prompts have the lineage limitation stated in
Section~\ref{sec:design}; they do not establish an official C4 test-set or
native attack-reproduction claim. Table~\ref{tab:cohort-roles} separates
confirmation from development analyses. Table~\ref{tab:math-access} specifies
the final editor's information; historical inference and stopping retain their
own observation contracts.

All 128 source reviews were available. Among the 23 excluded sources, nine
failed only minimum length, eight only paragraph completeness, and two only
the inherited-language criterion. Two failed both length and completeness,
one length and language, and one completeness and language. No source failed
all three conditions. The criteria were $N\ge40$, a completed paragraph and
$B\le1$; qualification did not use detector scores.

\begin{table}[ht]
\centering\small
\caption{Information boundary for the final construction. Public-model
surprisals are enabled only for the labeled weighted comparator.
Quality review and keyed scoring measure outcomes after selection.}
\label{tab:math-access}
\begin{tabular}{p{0.49\linewidth}cc}
\toprule
Object & Blind editor & Evaluator \\
\midrule
Source prose, pinned tokenizer, literal alternatives & Yes & Yes \\
Full-output $q,\ell,D,C$ & Yes & Yes \\
Public-model source surprisals, when enabled & Yes & Yes \\
True key, event states, source/output $Z,M$ & No & Yes \\
Original generation trace & No & Diagnostic only \\
Global source-relative quality judgment & After selection & Yes \\
\bottomrule
\end{tabular}
\end{table}

\begin{table}[ht]
\centering\small
\caption{Cohort roles and information access. Prompt/key counts denote
clusters, not additional observations. Development analyses are exploratory
and are not pooled with confirmation.}
\label{tab:cohort-roles}
\begin{tabularx}{\linewidth}{>{\raggedright\arraybackslash}p{0.20\linewidth}>{\raggedright\arraybackslash}p{0.21\linewidth}X}
\toprule
Study & Sources; clusters & Information and role \\
\midrule
Confirmation & 105 of 128; 16/8 & 82 initially detected; final coverage uses tokenizer-only structure. \\
Starting score & 95; 12/8 & Historical CUDA generation observations; aggregate inference. \\
Source nomination & 64; 8/8 & Historical MPS generation observations; candidate-supply diagnostic in the companion. \\
Ordinary rewriting & 16; 10/8 & Exposed development pairs; matched actual cost and mixed model reviewers. \\
Stopping & Same 105; 16/8 & Exposed bank; other-key training scores, without generation traces. \\
\bottomrule
\end{tabularx}
\end{table}

The source-model revision was
\texttt{989aa7980e4cf806f80c7fef2b1adb7bc71aa306}; the tokenizer JSON SHA-256 was\\
\texttt{c0382117ea329cdf097041132f6d735924b697924d6f6fc3945713e96ce87539}.
Generation and replay used the kernel in Section~\ref{sec:design}.
Detection retained original-CUDA partitions; CPU or MPS substitution would
define another evaluation. Provenance receipts retain input hashes and plotted
coordinates. The later hash-review overlay added 39 labels and preserved 66
existing labels without changing texts, scores or the primary hypothesis.

\subsection{Coverage solver agreement and local efficiency}
\label{app:solver-efficiency}

The bounded follow-up retained all 105 sources and 35,548 canonical outputs.
Both solvers recomputed global cost and net event counts from token paths;
cached metrics were used only for auditing. All candidate $q,D,C,T,\ell$ values,
1,521 attained-cost winners and 420 original allowance selections agreed.
Identical output hashes preserve the existing detector and quality outcomes;
this check adds no new efficacy observations.

Timing used Python 3.11.10 on arm64 macOS, one warmup and five alternating-order
repetitions per source. Inputs were already loaded; canonical-tokenization
checks, proposal generation and review were excluded. The reference used
bit-parallel edit distance and occurrence counters. Prefix sharing reduced
token-state transitions from 4,665,462 to 997,338, but added graph construction
and table-management work. Table~\ref{tab:solver-efficiency} distinguishes
cached solving from build-inclusive execution. Ratios above one favor DP.

\begin{table}[ht]
\centering\small
\caption{Coverage-DP efficiency on the materialized bank. Aggregate speedup
uses sums of per-source median times; it differs from a median paired ratio.}
\label{tab:solver-efficiency}
\begin{tabular}{lrr}
\toprule
Measurement & Cached graph & Build included \\
\midrule
Median paired speedup & $1.090\times$ & $0.878\times$ \\
Sources faster than enumeration & 61/105 & 38/105 \\
Sum of source median times (s) & 1.348 & 1.868 \\
Aggregate speedup & $1.678\times$ & $1.211\times$ \\
\bottomrule
\end{tabular}
\end{table}

Enumeration's summed source-median time was 2.262 seconds. DP reduced aggregate
workload time while slowing the median source when construction was included.
These conditional measurements differ from the full local pipeline timing in
Section~\ref{sec:construction}. On eight prospectively size-stratified sources,
median incremental Python allocation peaks were 48.6 KiB for enumeration,
69.9 KiB for cached DP solving and 918.2 KiB for build-plus-DP; the latter peaked
at 43.45 MiB. Cached solving excludes retained graph memory, and these
allocations are not process RSS. Runtime repetitions characterize this local
implementation, without increasing the statistical sample for evasion claims.

\subsection{Candidate nomination and mechanical admission}
\label{app:construction-details}

The fixed nomination rule split each source at punctuation and retained units
with at least six words. A unit longer than eighteen words was clipped to its
first twelve words. If more than eight units remained, spatially distributed
quantiles were selected in source order. The resulting disjoint character spans
need not be complete syntactic clauses; trailing punctuation stayed in the
source. Although historical metadata included density ranks, the standalone
constructor reproduced its requests with zero-valued source features.

Admission preserved named lexical anchor counts, quantities, written numbers,
modal and negation words, temporal markers and bracket counts. It rejected
new non-ASCII letters and nonordinary control characters and required a changed
replacement. The single-site complete output needed global token distance
between one and eight, net length change at most $0.05N$, at most sixty-four
replacement tokens and exact encode/decode round-trip. Missing or malformed
responses remained failures without automatic retries. Global semantic
preservation was measured separately after composition and selection.

For $s\le8$ sites with $m_j\le2$ admitted alternatives each, the literal language is
\begin{equation}
  \mathcal{L}(x)=
  \left\{\operatorname{Compose}(x,a_1,\ldots,a_s):
    a_j\in\{\mathrm{keep},1,\ldots,m_j\}\right\}.
  \label{eq:construction-language}
\end{equation}
Here $x$ denotes the source path together with its prose. Complete compositions
undergo the canonical round-trip and resource checks in
Eq.~\eqref{eq:construction-feasible}, even when their individual alternatives
passed admission. This captures retokenization at replacement boundaries.

\subsection{Probability comparator and constant-weight control}
\label{app:selection-comparators}

The weighted comparator used the same language and actual-cost slices. Let
$s_t$ be public-model source-token surprisal and
$\widehat F(s)=T^{-1}\sum_{t=2}^{N}\mathbf{1}\{s_t\leq s\}$.
Its source-event utility was
\begin{equation}
  w(e)=\frac{1}{n_x(e)}
    \sum_{\substack{2\leq t\leq N\\(x_{t-1},x_t)=e}}
      \frac{\widehat F(s_t)}{2},
  \qquad e\text{ occurring in }x.
  \label{eq:construction-rank-weight}
\end{equation}
Absent-source identities received zero utility. The rule minimized
\begin{equation}
  \widehat Z_{\mathrm{rank}}(x')=
    \frac{\sum_e n_{x'}(e)w(e)}{0.5\sqrt{T'}}
  \label{eq:construction-rank-proxy}
\end{equation}
at $q=q_\rho$, followed by site-count, candidate-ID and text-hash ties.
This rank utility has no established calibration as posterior centered evidence;
zero utility does not infer a red state. Its normalization, aggregation and
ties also differ from coverage, so the primary comparison changes complete
selection policies.

The retrospective constant-weight control isolated the source weights while
preserving that denominator and tie rule. Assigning positive constant $c$ to
source identities and defining
$C_{\mathrm{seen}}=\sum_{e:n_x(e)>0}[n_{x'}(e)-n_x(e)]_+$ gives
\begin{equation}
\widehat Z_c(x')=
\frac{2c\,[T-D+C_{\mathrm{seen}}]}{\sqrt{T'}}.
\label{eq:construction-constant-control}
\end{equation}
This counts uncapped output occurrences of source identities. Its difference
from coverage can therefore reflect length, surplus occurrences and ties.

\subsection{Quality measurement and exact-identity reuse}

The frozen source gate and relative/polished thresholds are specified in
Section~\ref{sec:design}. Appendix~\ref{app:execution} reproduces the executed
ABC-v2 instruction, including every score level. Reviewer requests contained
anonymous prose pairs and no keys, states, scores or policy labels. Exact
source/output identities alone permitted reuse. Unreviewed outputs remained
unavailable until an explicitly recorded completion.

Ten controls tested the policy: identity, paraphrase, inherited defect, benign
emphasis and small readable fluency damage were accepted; quantity, negation,
role, modality and causation changes were rejected. All ten decisions matched
expectation. These checks establish instruction compliance for the controls,
without human agreement or independent factual validation. The absolute level
adds $B=0$ to polished quality; all quality levels remain source-relative.

\subsection{Clustered uncertainty}
\label{app:bootstrap}

Confirmation intervals use 10,000 crossed prompt/key bootstrap draws with
seed 130913. Each draw independently resamples the observed prompt and key
labels with replacement, retaining the number of clusters on each axis.
A document receives the product of its sampled prompt and key multiplicities.
This is the pigeonhole resampling construction \citep{owen2007pigeonhole}.
Weighted means use document outcomes or within-document policy differences,
with the same draws across compared outcomes and doses. Draws with zero total
weight are omitted and their number is recorded; they are not replaced.
The 2.5th and 97.5th percentiles of the remaining draws give pointwise 95\%
intervals. All 10,000 draws were nonempty for the displayed confirmation
dose curves. The ordinary-rewriting comparison uses its own clusters and
effective draw counts (Appendix~\ref{app:ordinary-comparison}). Sixteen prompt
and eight key clusters limit confirmation precision; candidates, repeated
events and doses do not enlarge the independent sample.

\subsection{Finite-bank and follow-up protocols}

The retrospective audit enumerated 37,036 configurations and retained 35,548
eligible unique outputs, reproducing canonical IDs, costs, incidence, hashes
and all 1,680 original endpoints. Blind controls and structural subgroups
were sealed before additional actual-key scoring. Original-CUDA scoring loaded
no language model. Previously examined sources retain exploratory status.

At each original actual-cost slice, uniform choice was its exact arithmetic
expectation over unique outputs, and the evaluator oracle minimized realized
$Z'$. Singleton or all-tied slices received tie-aware percentile $0.5$.
Choice-rich slices had at least five outputs; variable-$D$ slices were specified
before scoring. At-most coverage maximized $D$ across all costs under the cap,
using lower $q$ for coverage ties. Source summaries preceded crossed prompt/key
bootstrap inference; candidate counts did not enlarge the sample.

Minimum-cost witness review and the separate stopping protocol appear in
Appendix~\ref{app:stopping}. Those analyses reused exposed outcomes; the
stopping fits excluded held-out prompt and key groups and used other-key
training scores. No quality labels or test-key states entered their selection.

\subsection{Standalone replay and resource scope}

A standalone worker with the pinned Rust tokenizer replayed sanitized source
text and recorded proposal responses. It reproduced all 1,018 request bodies,
803 admitted patches and 420 coverage outputs on 105 sources, with zero
archived-run reads, heavy-model imports, network calls or loaded model weights.
Five timing passes supplied 2,100 identical output checks, without new efficacy
observations. Tokenizer loading took $91.949$ ms; selection median and 95th
percentile were $8.917$ and $335.475$ ms. Absolute peak RSS was $108.2$ MiB,
without subtracting interpreter memory. These one-host measurements exclude
provider proposals and review inference. Comparative end-to-end speed remains
unmeasured.

%% file: sections/appendix_reporting_controls.tex
\section{Complete outcomes, controls and measured examples}
\label{app:additional-outcomes}

\subsection{Outcome definitions and dose summaries}
\label{app:dose-metrics}

For policy output $x'_i(\rho)$ at allowance $\rho$, let $\mathcal P$ contain
qualified sources with $Z_i(0)>\tau$ and let $n_+=|\mathcal P|$. Define
\begin{align}
c_i(\rho)&=\frac{q(x_i,x'_i(\rho))}{N_i},\qquad
\bar c_+(\rho)=\frac1{n_+}\sum_{i\in\mathcal P}c_i(\rho),\nonumber\\
R_+(\rho)&=\frac1{n_+}\sum_{i\in\mathcal P}
  \ind\{Z_i(\rho)<\tau\}.
\label{eq:metrics-cost-raw}
\end{align}
For relative and polished quality indicators $Q_i$ and $P_i$, joint rates and
continuous suppression are
\begin{align}
J_+(\rho)&=\frac1{n_+}\sum_{i\in\mathcal P}
  \ind\{Z_i(\rho)<\tau\}Q_i(\rho),\nonumber\\
J_{+,\mathrm{pol}}(\rho)&=\frac1{n_+}\sum_{i\in\mathcal P}
  \ind\{Z_i(\rho)<\tau\}P_i(\rho),\nonumber\\
S_+(\rho)&=\frac1{n_+}\sum_{i\in\mathcal P}
  \bigl[Z_i(0)-Z_i(\rho)\bigr].
\label{eq:metrics-quality-reduction}
\end{align}
Quality failure leaves the denominator unchanged. For the same selected
outputs, $J_{+,\mathrm{pol}}\le J_+\le R_+$; this does not bound other
candidates' quality-qualified success. Negative reductions remain in $S_+$.
Mean actual cost is an operating point under the cap, not a shared individual
cost. Table~\ref{tab:coverage-dose-metrics} uses the same 82 positives for
both coordinates.

\begin{table}[t]
\centering\small
\caption{Dose outcomes on 82 initially detected qualified sources. Cost is
mean actual $100q/N$; escape columns are strict counts out of 82. Intervals
are pointwise crossed prompt/key 95\% intervals for mean $\Delta Z$ (16/eight
clusters). Four allowances use separately selected, nonnested outputs.
Coverage includes retrospective completion of its 25 missing crossing reviews.}
\label{tab:coverage-dose-metrics}
\label{tab:dose-metrics}
\begin{tabular}{lrrrrrrr}
\toprule
Policy & Cap & Cost & Raw & Relative & Polished & $\Delta Z$ & 95\% CI \\
\midrule
Coverage & 5\% & 4.33\% & 13 & 13 & 13 & 0.436 & [0.293, 0.581] \\
 & 10\% & 9.31\% & 25 & 24 & 22 & 0.930 & [0.716, 1.142] \\
 & 15\% & 13.41\% & 39 & 37 & 32 & 1.255 & [1.012, 1.515] \\
 & 20\% & 15.96\% & 40 & 39 & 34 & 1.373 & [1.089, 1.660] \\
\midrule
Weighted & 5\% & 4.33\% & 16 & 16 & 15 & 0.496 & [0.360, 0.634] \\
 & 10\% & 9.31\% & 29 & 27 & 24 & 0.977 & [0.791, 1.171] \\
 & 15\% & 13.41\% & 39 & 37 & 31 & 1.228 & [0.986, 1.465] \\
 & 20\% & 15.96\% & 40 & 39 & 34 & 1.359 & [1.066, 1.649] \\
\bottomrule
\end{tabular}
\end{table}

All 420 weighted outputs have reviews. Coverage's 25 formerly unreviewed
crossings comprised 3, 10 and 12 distinct pairs at the three lower allowances.
The frozen pairs received the unchanged Sol-medium ABC-v2 rubric after the
ten controls passed: all 25 passed relative quality and 24 were polished.
Existing labels, texts and scores were unchanged. Only exact identities gained
labels. Every coverage crossing is now reviewed; some noncrossing lower-dose
outputs remain unreviewed, leaving overall quality-pass rates incomplete.

The original all-105 weighted mean costs were $4.12/9.20/13.38/16.04\%$,
with mean $\Delta Z=0.441/0.930/1.184/1.333$. These include 23 initially
nonpositive sources. At p20, mean cost was $14.50$ tokens; counting excluded
initially positive attempts as operational nonsuccesses gives 39/91, compared
with 39/82 qualified positives. Coverage and weighted successes overlap on
38 sources, with coverage-only \texttt{s088} and weighted-only \texttt{s062}.
Their success-conditioned mean costs are $17.31\%$ and $17.32\%$.

From p15 to p20, coverage gained five qualified escapes and lost three;
weighted gained four and lost two. Scores worsened for 16/82 sources under
each policy and for 21/105 qualified sources under weighted selection.
The p20 positive-subgroup actual-cost interquartile range was
$13.93$--$19.22\%$, with mean $15.96\%$. Neither an aggregate curve nor its
mean cost establishes monotonic improvement or a common realized edit count.

\subsection{Selection controls and finite-bank opportunity}
\label{app:selection-controls}

\begin{table}[t]
\centering
\caption{Fully reviewed $20\%$-allowance selections on the same candidate bank and identical actual-cost slices. Mean $\Delta Z$ and relative-quality passes use all 105 qualified sources; escape columns use the 82 initially positive qualified sources. Joint means strict $Z'<4$ and relative quality. Polished joint additionally requires $F=0$. The hash quality column includes retrospective completion of 39 labels.}
\label{tab:policy-results}
\begin{tabular}{lrrrrr}
\toprule
Policy & Mean $\Delta Z$ & Relative /105 & Raw /82 & Joint /82 & Polished /82 \\
\midrule
Weighted & 1.333 & 101 & 40 & 39 & 34 \\
Coverage & 1.313 & 100 & 40 & 39 & 34 \\
Fixed hash & 1.305 & 102 & 38 & 37 & 32 \\
3B weighting & 1.334 & 101 & 39 & 38 & 33 \\
\bottomrule
\end{tabular}
\end{table}

Table~\ref{tab:policy-results} preserves the complete largest-dose outcomes.
The weighted 39/82 rate has crossed 95\% interval $[27.8\%,67.3\%]$;
the corresponding coverage interval is $[28.6\%,66.7\%]$.
Among the 82 positive sources, 26 attained-cost menus contained one unique
output, 22 varied in destroyed-event count, and six contained both crossing
and noncrossing outputs. These counts describe the available selection
opportunity; they do not establish equivalence between selectors.

The primary complete-rule comparison is in Section~\ref{sec:results}.
Table~\ref{tab:bank-audit} separates the retrospective weight ablation,
uniform-choice comparison and actual-key oracle. They use the fixed protocol
in Appendix~\ref{sec:appendix-evidence} and the same attained-cost slices.

\begin{table}[t]
\centering\small
\caption{Retrospective mean extra $\Delta Z$ with pointwise crossed 95\%
intervals across 105 source-paired observations. Uniform is exact within-source
expectation over unique outputs; the minimum-$Z'$ oracle uses hidden states.
Candidate rows and doses do not increase the independent source count.}
\label{tab:bank-audit}
\begin{tabular}{lrrr}
\toprule
Allowance & Weighted $-$ constant & Coverage $-$ uniform & Oracle $-$ coverage \\
\midrule
5\% & $0.058\ [-0.014,0.144]$ & $0.019\ [-0.065,0.103]$ & $0.288\ [0.183,0.410]$ \\
10\% & $0.063\ [-0.027,0.168]$ & $0.056\ [-0.028,0.151]$ & $0.365\ [0.253,0.488]$ \\
15\% & $-0.0003\ [-0.091,0.082]$ & $0.087\ [0.004,0.175]$ & $0.191\ [0.107,0.311]$ \\
20\% & $0.013\ [-0.054,0.087]$ & $0.019\ [-0.035,0.072]$ & $0.139\ [0.077,0.223]$ \\
\bottomrule
\end{tabular}
\end{table}

At p20, constant-minus-coverage added $0.0066$ $[-0.0039,0.0340]$;
together with the weight contrast it reproduces the original $0.0196$
policy difference. Constant weights also gave 40 raw and 39 reviewed joint
escapes. Coverage-minus-fixed-hash gain was $0.0086$ $[-0.0655,0.0876]$,
with joint difference $+2.44$ percentage points $[-5.13,+10.98]$.
The hash control uses one deterministic ordering. These secondary comparisons
establish neither equivalence nor separate effects of the shared proposal,
admission and dose stages.

Restricted candidate choice limits what these comparisons can identify.
The p15 pointwise gain in Table~\ref{tab:bank-audit} is exploratory among
several doses; small choice-rich subgroups do not replace the primary null.
The technical companion
retains menu sizes, selection turnover, uniform-choice percentiles and
subgroup intervals.

Let $\mathcal E_\rho(x)=\{u\in\mathcal F_\rho(x):q(x,u)=q_\rho\}$.
For the realized key, nested feasibility gives
\begin{equation}
\min_{u\in\mathcal F_\rho(x)} Z_K(u)
\;\le\;\min_{u\in\mathcal E_\rho(x)} Z_K(u)
\;\le\; Z_K(x'_\rho).
\label{eq:bank-oracle-ordering}
\end{equation}
Identity ensures nonempty feasibility. Same-cost oracle raw counts were
23/40/47/42 of 82 across doses, while all-cost counts were 26/49/53/56.
Only all-cost capacity must be nondecreasing. No quality pass is imputed
to unreviewed oracle outputs. Blind at-most coverage gave 41/82 raw escapes,
mean $q=14.15$ versus $14.50$, and extra mean $\Delta Z=0.0372$
$[-0.0328,0.1232]$. Its current qualified count remains 36--40: one known
failure and four exact outputs unreviewed even after later completions.

\subsection{Objective correction and ordinary-rewriting comparison}
\label{app:ordinary-comparison}

The all-cost reset rule minimized $[T-D+C_{\mathrm{seen}}]/\sqrt{T'}$
with identity fallback in the same 105-source bank. At p20 its extra mean
$\Delta Z$ over at-most coverage was $0.007025$ $[-0.034336,0.061242]$,
whose upper endpoint is below the tested $0.1$-Z reference. This retrospective
null concerns that objective and bank, without ruling out other estimators.

The ordinary-rewriting pilot retained 16 exposed development sources and
the frozen weighted reference with its sealed proposal-depth repair. Both
arms had mean actual $q=15.125$ and cost $17.222\%$. Fourteen pairs met the
pre-existing combined dose/quality gate; all 16 remained in analysis. All
selected outputs passed relative review. The reference used eleven cached
Sol and five new Astra judgments, and the baseline used sixteen new Astra
judgments under the same rubric. This was neither common-reviewer nor human
confirmation.

The baseline requested sixteen independent literal-edit plans per source from
Luna at low reasoning, capped at eight self-nominated disjoint edits and
1,024 output tokens per call. It assembled coherent plans through sixteen
fixed hash-priority greedy restarts per cap. Selection used actual cost and
output hashes, with no $D$, source probabilities, quality labels or scores.
Both arms used the largest positive actual cost available to both under the
allowance, returning identity and retaining the source if none existed.
Canonical round-trip, global cost, ordinary formatting and $5\%$ length
constraints applied, without the reference's specialized factual-anchor guard.
Of 256 primary baseline calls, 207 were valid and 49 failed; the reference
used 198 historical/repair calls. Matching final cost did not match supply
or inference expenditure.

The extra mean $\Delta Z$ was $0.010634$ $[-0.486268,0.508954]$ over ten
prompt/eight key clusters. Each arm escaped on 7/10 positives: six shared,
one unique to each and two under neither. Mean margin difference was
$0.102729=0.250000-0.187500+0.040229$, separating destroyed, created and
length contributions. Only sealed primary matched outputs were scored.
Of 10,000 bootstrap draws, 9,997 mean-effect and 9,965 positive-crossing
resamples were nonempty. Unequal supply, mixed reviewers, source reuse and
the wide interval leave superiority and equivalence unresolved; this was
not a native SIRA/BIRA comparison.

\subsection{Signed conversion, initial headroom and repair}
\label{app:reset-residual}
\label{app:witness-diagnostics}

For the frozen weighted outputs at the largest dose, the mean decomposition
across all 105 qualified sources is
\begin{equation}
\overline{\Delta M}
=\overline{A_D}+\overline{-A_C}+\overline{\Delta b}
\simeq6.142857+0.204762-0.035875
=6.311744.
\label{eq:observed-margin-decomposition}
\end{equation}
The 2,230 destroyed occurrences had pooled green rate $78.92\%$, and the
2,197 created occurrences had rate $49.02\%$. These correlated occurrences
are descriptive evaluator evidence, not independent inferential units.

The main witnesses illustrate why source strength alone cannot predict a
rewrite. For analyst reference $\epsilon_0=A_0/T$, their reset residual is
\begin{align}
L-\epsilon_0D&=(A_D-\epsilon_0D)-A_C,\nonumber\\
\Delta Z-\Delta Z_{\mathrm{reset}}
 &=\frac{A_D-\epsilon_0D-A_C}{\sigma\sqrt{T'}},\nonumber\\
\Delta Z_{\mathrm{reset}}
 &=Z_0\left[1-\left(1-\frac DT\right)\sqrt{T/T'}\right].
\label{eq:reset-residual}
\end{align}
Table~\ref{tab:reset-witnesses} retains both source-yield and created-state
departures. These are different-source, retrospectively chosen illustrations,
not a matched causal comparison or blind prediction. Both reset-reference
scores remain above threshold, while the realized signed effects separate
escape from detection. Figure~\ref{fig:mechanism-examples} gives their complete
margin accounting; its numerical inputs are also retained in the
technical companion.

\begin{table}[t]
\centering
\caption{Evaluator reset diagnostics for the two main-text witnesses. Both
complete outputs passed relative and polished review; source, cost and key
state differ. The homogeneous reference uses actual source evidence.}
\label{tab:reset-witnesses}
\begin{tabular}{lrr}
\toprule
Quantity & \texttt{s029} & \texttt{s001} \\
\midrule
Source/output scored events $T,T'$ & $71,69$ & $132,128$ \\
Source centered evidence $A_0$ & $26.5$ & $33$ \\
Destroyed-event count $D$ & $21$ & $32$ \\
Homogeneous source reference $\epsilon_0D$ & $7.838$ & $8.000$ \\
Selected-source excess $A_D-\epsilon_0D$ & $+0.662$ & $-2.000$ \\
Created contribution $-A_C$ & $+1.500$ & $-3.000$ \\
Actual evidence reduction $L$ & $10.000$ & $3.000$ \\
Final $Z$ under reset reference & $4.493$ & $4.419$ \\
Measured final $Z$ & $3.973$ & $5.303$ \\
\bottomrule
\end{tabular}
\end{table}

The first output replaced $37.74\%$ of centered source evidence against a
$37.31\%$ requirement; the second replaced $9.09\%$ against $31.43\%$.
Figure~\ref{fig:decomposition} retains all sources, including negative gains.
The budget floor excluded 20/82 at the allowed $5\%$ dose and 22 at its
attained cost, and none at the three larger doses. Its median was 3.5 edits;
median attained p20 cost exceeded the individual floor by twelve. Passing
this necessary bound was therefore permissive at larger allowances.

\input{figures/support_panels}

Among the 39 weighted joint successes, the two-sided certificate in
Appendix~\ref{app:math-stability} certified one additional token edit for 32,
two for 18 and three for seven. Seven had radius zero, including the opening
example. Median successful $Z'$ was $3.151$; 29 were below $3.5$ and fifteen
below three. These evaluator-side certificates concern detector decisions,
without semantic guarantees.

\subsection{Quality examples and the grounding boundary}

One raw escape changed ``decision on dropping charges'' to ``decision to drop
charges.'' The reviewer marked introduced risk: an unresolved question could
become an apparently settled decision. Its score fell $5.683\to3.674$, but
it failed joint quality. The accepted complete witness below includes
``exposed grave misconduct'' becoming ``revealed serious misconduct.'' These
are complete-output judgments; individual displayed phrases have no independent
quality or causal score attribution.

The grounding pilot used context-bearing records from Dolly
\citep{databricks2023dolly}. Our generated-source cohort yielded zero
grounded sources, leaving grounded-source-conditioned escape undefined.
Its historical Q3 labels are separate from confirmation ABC-v2. The
technical companion
retains attempt counts and detector outcomes; these do not establish
transfer to grounded-source rewriting.

\subsection{Complete source and constructed output}
\label{app:complete-example}

Both coverage and weighted p20 selected the following \texttt{s029} output.
Bold intervals are the three source-character replacements, including retained
context. Global cost was $q=14$ for $N=72$, with $\ell=-2,D=21,C=19$.

\paragraph{Source, $Z=6.290$.}
\begin{quote}\small
\textbf{After two weeks following the establishment of a Special Committee tasked with}
investigating violent and unruly events in the Canadian parliament in November
last month, \textbf{their comprehensive report exposed grave misconduct that
deeply disturbed the democratic process}, highlighting instances of intimidation,
disorderly conduct, and blatant disregard for protocol, \textbf{thereby prompting
calls for enhanced regulations to ensure parliamentary conduct adheres to high
ethical standards}.
\end{quote}

\paragraph{Output, $Z'=3.973$.}
\begin{quote}\small
\textbf{Two weeks after the establishment of a Special Committee tasked with}
investigating violent and unruly events in the Canadian parliament in November
last month, \textbf{their comprehensive report revealed serious misconduct that
profoundly disrupted the democratic process}, highlighting instances of
intimidation, disorderly conduct, and blatant disregard for protocol,
\textbf{thus prompting demands for stronger regulations to ensure that
parliamentary conduct meets high ethical standards}.
\end{quote}
The reviewer passed relative and polished quality. The inherited awkward date
expression remains; historical factual accuracy was unmeasured. Terminal margin
was approximately $-0.113$, giving sufficient repair radius zero despite the
larger score reduction.

%% file: figures/support_panels.tex
\begin{figure}[t]
  \centering
  \begin{subfigure}{0.49\linewidth}
    \includegraphics[width=\linewidth]{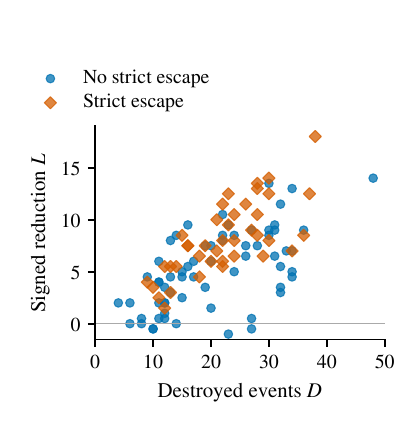}
    \caption{Support and signed effect, all 105 sources.}
  \end{subfigure}\hfill
  \begin{subfigure}{0.49\linewidth}
    \includegraphics[width=\linewidth]{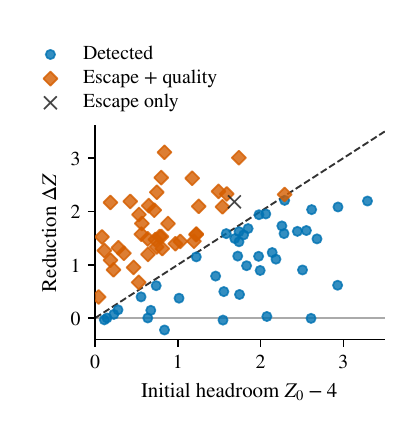}
    \caption{Crossing initial headroom, 82 positives.}
  \end{subfigure}
  \caption{Evaluator-side variation in frozen weighted p20 selections. Panel
  (a) uses all 105 qualified sources and distinguishes raw strict escape from
  other outcomes; destroyed-event count does not uniquely determine signed
  reduction. Panel (b) uses 82 initially positive sources and places achieved
  $\Delta Z$ against required headroom $Z_0-4$. Strict escapes lie above the
  dashed equality line: 39 passed relative quality and one did not. Blue circles
  remained at or above four. Measured evaluator scores are not editor inputs.
  All observations, including negative gains, are retained; event occurrences
  and candidate outputs are not independent inference units.}
  \label{fig:decomposition}
\end{figure}

%% file: sections/appendix_historical.tex
\section{Initial-strength estimation and information limits}
\label{app:historical-information}

\subsection{Historical cohort and approximate estimator}

The 95-source cohort used original-CUDA generation and scoring over twelve
prompt/eight key clusters. It first supplied an independent supply-expansion
replication and was subsequently reused for strength estimation, allocation
and trace ablation. Those later analyses are exposed-cohort diagnostics,
separate from the 105 qualified confirmation sources. The planned 96th cell,
\texttt{tc4\_p8\_k7}, was excluded during site preparation because the saved
generation had \texttt{retokenization\_exact=false}. The pre-existing canonical
check preceded strength estimation; exclusion did not depend on its error.

Starting-strength inference used stored generation-time pre-bias logits and
known sampling processors. Context-row inference retained 400 scenarios with
a base importance-sampling batch parameter of 3,000. Difficult rows used
1,500 auxiliary prior draws, further base-sized proposal batches and, when
needed, a Gibbs-informed batch of 6,000. MCMC or prior fallbacks handled
unresolved rows, so 3,000 was not a fixed total sample count per row. Stored
logits were truncated and the partition prior was approximate. The implementation
formed centered source estimate $C_0$ and stored
$\widehat Z_0=2C_0/\sqrt T$ for $\gamma=1/2$. Rechecking all 95 saved rows
reproduced this arithmetic exactly, without regenerating detector states.

The main constructor uses neither these observations nor the estimator.
An earlier 20-document source-state study and the historical nomination,
supply and prefix diagnostics are retained in the technical companion with
their separate access and quality limitations.

\subsection{Aggregate association and uncertainty}

The measured correlation was $0.792855$, mean estimate-minus-measurement error
$0.089280$, and population error standard deviation $0.769175$. The crossed
prompt/key interval for mean error was $[-0.241,0.378]$. An in-sample regression
on an intercept and $\sqrt T$ had $R^2=0.493557$; adding $\widehat Z_0$ gave
$0.629916$, with partial correlation $0.518891$ after length residualization.
Figure~\ref{fig:starting-strength} displays these unchanged source rows.
This descriptive adjustment establishes neither unbiasedness nor held-out
calibration, plaintext inference or improved allocation beyond length.

For the ideal conditional model, source-count vector $n_0$ and covariance
$\Sigma=\operatorname{Cov}(y\mid\mathcal I)$ give
\begin{equation}
\operatorname{Var}(Z_0\mid\mathcal I)
=\frac{n_0^\top\Sigma n_0}{\sigma^2T}.
\label{eq:starting-score-variance}
\end{equation}
Weak enough dependence can yield $O(1)$ score uncertainty when source evidence
variance is $O(T)$, while stable positive enrichment gives mean score
proportional to $\sqrt T$. Repeated identities and partition conditioning
affect covariance; a diagonal approximation is not a general lower bound.
The exchangeable working prior differs from the active finite modulo-reseed
implementation. Thus Eq.~\eqref{eq:starting-score-estimate} defines an ideal
mean but only an operational approximation for the implemented estimator.
The aggregate and candidate-contrast directions in
Appendix~\ref{app:math-uncertainty} explain why this association does not
validate a replacement ranker.

\subsection{Allocation control and interpretation}

To distinguish strength-informed allocation from concentration on short
sources, the decisive control ordered saved 30\%-dose outputs under the recorded cap:
length-first and estimate-first each yielded 32 crossings, versus 33 for
triage at slightly different expenditures. Random concentration averaged
about 20.8. These outcomes identify concentration and length as alternative
explanations and leave added Z0-based allocation value unresolved. The historical
reviewer's identity gate failed, so these are raw descriptive outcomes, not
quality-qualified or uniform-20\% confirmation rates. The
technical companion (Section 3) retains
the preliminary allocation sequence, expenditure differences, rules and
prefix plots. A temporary low prefix does not imply
terminal escape, and full-trace conditional prefix estimates are not online
forecasts.

\subsection{Reconstructed same-bank source-evidence selection}
\label{app:reconstructed-selection}

The historical estimates did not match the confirmation sources. A subsequent
repair replayed preserved prompts and generated token paths using the pinned
float32 CUDA generator. On 238 archived historical rows, replay reproduced all
stored top-1500 logit values exactly and the top-20 sets agreed. Closing-source
replays were repeatable, but their original logits were not archived, so direct
historical equality was untestable there. This diagnostic requires generator
access beyond the confirmed constructor's information contract.

The historical approximate inference procedure retained 400 scenarios, with
base importance batch parameter 3,000. It resolved 104 sources; one canonical
versus generated token-path mismatch triggered coverage fallback. All 7,855
context records had positive importance-sampling effective size; the minimum
was 173.20. These computational checks do not establish posterior calibration.
The independent-context approximation and historical observation handling remain.

Before reading keyed outcomes, the selector maximized
$\widehat W(x')=\sum_e d_e\widehat m_e$ at each coverage reference's exact actual
cost, retaining coverage's deterministic ties. Two secondary controls also
fixed output length, or fixed $q,D,C,T'$. The candidate bank and enumeration
solver stayed unchanged; the tested utility estimates destroyed source evidence
only. The primary was the resolved-source contrast at the 20\% allowance.

\begin{table}[ht]
\centering\small
\caption{Reconstructed evidence selection minus coverage at matched actual
cost. Intervals are crossed prompt/key 95\% intervals; controls are secondary.}
\label{tab:reconstructed-selection}
\begin{tabular}{lrrl}
\toprule
Comparison & $n$ & Extra $\Delta Z$ & 95\% interval \\
\midrule
Resolved, primary & 104 & 0.0314 & $[-0.0372,0.0988]$ \\
Operational, with fallback & 105 & 0.0311 & $[-0.0369,0.0981]$ \\
Initially detected & 82 & 0.0070 & $[-0.0672,0.0667]$ \\
Same output length & 104 & 0.0117 & $[-0.0603,0.0728]$ \\
Same $D,C,T'$ & 104 & 0.0209 & $[-0.0269,0.0772]$ \\
\bottomrule
\end{tabular}
\end{table}

The primary changed 22 outputs: 13 improved $Z$, six worsened it and three tied.
Both policies produced 40 raw crossings among 82 positives, with different
crossing identities. The inferred selector had 37 reviewed relative-quality
passes, one failure and two unreviewed crossings, giving 37--39 possible qualified
crossings against coverage's 39. Missing reviews remain unknown. The intervals
use 10,000 crossed replicates over sixteen prompt/eight key clusters and condition
on the sealed estimates; they exclude inference-resampling uncertainty.

The exact extra mean margin reduction decomposed as
$0.124780=0.038462+0.081731+0.004587$: destroyed-source evidence, favorable created
evidence and the length boundary. Every component's interval included zero.
The source-only selector did not model created-state effects. A descriptive
reconstructed $Z_0$ check gave correlation 0.82845, mean error $-0.02027$ and
RMSE 0.65207 on 104 sources. This reused-cohort diagnostic and the uncertain
selection increment distinguish aggregate estimation from candidate contrasts;
they establish neither held-out stopping calibration nor a new quality advantage.

%% file: sections/appendix_stopping.tex
\section{Reviewed minimum costs and key-blind stopping}
\label{app:stopping}

\subsection{Reviewing the minimum-cost witnesses}

The retrospective minimum-cost audit used the 35,548 previously scored outputs
from the 105 qualified sources. For each of the 56 initially detected sources
with a crossing in the allowed bank, we identified the smallest actual token
cost admitting $Z'<4$. Eight witnesses reused passing exact-identity reviews;
48 deterministic representatives received blinded GPT-6.1 Sol-medium review
under the unchanged rubric after all ten control pairs passed. The reviewer
received prose and preservation instructions, without detector scores, colors
or selection status. Witness discovery itself used actual-key outcomes.

All 56 witnesses passed relative quality; 51 had $F=0$ and five had
$F=1$; all had $C_{\mathrm{risk}}=0$. Their polished status describes these
selected witnesses and does not determine the minimum cost of every polished
alternative in the bank. The quality-constrained minimum equals the raw
minimum because a passing output attains the raw lower bound. This is a
finite-bank certificate under the recorded model rubric. Other oracle outputs
retain their own review status, and these retrospectively discovered witnesses
do not constitute an operational attack rate.

Conditional on these 56 sources, median minimum cost was $5.70\%$.
This conditional median is distinct from the same-source mean-cost comparison
on 39 coverage successes in Section~\ref{sec:results}.

\subsection{Frozen paths, prediction and stopping}

The stopping study retained all 105 sources and the existing candidate bank,
with $q\le\lfloor0.2N\rfloor$ and $|\ell|\le0.05N$. One candidate at each
attainable actual cost formed a coverage path, selected by maximal $D$ and
the original deterministic ties. A secondary path used the existing public-model
surprisal utility. Identity outputs were included. Paths, features and rules
were frozen before comparison, but the bank and evaluator outcomes had already
been examined. This was a retrospective reuse of existing outputs and scores.

For candidate $j$, let $C_{\mathrm{seen},j}$ count created occurrences whose
event identity also occurs in the canonical source. A homogeneous-source
reference predicts remaining evidence through
\begin{equation}
R_j=T-D_j+C_{\mathrm{seen},j},\qquad
a_j=\frac{R_j}{\sqrt{TT'_j}},\qquad
\widehat Z_j=a_j\widehat Z_0.
\label{eq:stopping-predictor}
\end{equation}
The length model S0 fitted a common enrichment $\widehat\epsilon$, giving
$\widehat Z_0=2\widehat\epsilon\sqrt T$. Model S1 used
$\widehat\epsilon_i=\operatorname{logistic}(\beta_0+\beta_1h_i)-1/2$, where
$h_i$ is mean cached public-model source surprisal standardized using training
sources. Fitting minimized source-balanced squared baseline-score error, with
a fixed $0.01\beta_1^2$ penalty for S1. This scalar plaintext feature differs
from the historical trace-assisted estimator in
Section~\ref{sec:historical-information}.

Four prompt groups crossed with four key groups defined 16 outer folds.
Training excluded both held-out groups. The empirical 90th percentile of
inner held-out source-wise maximum residuals supplied a nonnegative correction
$c$ for each outer fold and path type. The rule selected the
cheapest path output with $\widehat Z_j+c<z_\star$, for prespecified targets
$z_\star\in\{4,3.8,3.5\}$. Absence of such an output returned the original
coverage-p20 output as unassured best effort. Mean-only controls set $c=0$.
A fixed-dose control chose the cheapest training allowance whose raw yield
was within five percentage points of training p20 yield; every outer fold
selected $15\%$.

These controllers used detector scores from other training prompt/key groups.
They were test-key-blind, while the original tokenizer-only constructor required
no such score training. Quality labels and test-source detection status were
excluded from selection. The empirical envelope supplies no distribution-free
or per-declaration success guarantee. Whole-path coverage, declaration frequency
and success among declared outputs are separate measurements.

\subsection{Starting-score and signed-effect errors}

The prediction error can be separated exactly using evaluator-side
$\epsilon_0=A_0/T$ and $L_j=A_{D,j}-A_{C,j}$:
\begin{equation}
Z_j-\widehat Z_j
=\underbrace{a_j(Z_0-\widehat Z_0)}_{U_j}
+\underbrace{\frac{\epsilon_0(D_j-C_{\mathrm{seen},j})-L_j}
{\sigma\sqrt{T'_j}}}_{V_j}.
\label{eq:stopping-error}
\end{equation}
Here $U_j$ propagates starting-score error, while $V_j$ remains after replacing
the estimate with the true starting score. It includes source heterogeneity,
created and reused event states, and error in the homogeneous-reset reference.
It is not an estimate of irreducible cryptographic noise. The components are
correlated, requiring the cross term in
$\mathbb E[(Z_j-\widehat Z_j)^2]=\mathbb E[U_j^2]+\mathbb E[V_j^2]+2\mathbb E[U_jV_j]$.

Across all 105 sources, baseline RMSE was $0.792$ for S0 and $0.780$ for S1;
S1's mean coverage-path correction was $1.41Z$. The empirical envelope covered
the complete path for 96/105 sources and 74/82 positives; this is path coverage,
distinct from correctness conditional on declaring a target attained.
These small predictive differences did not produce a useful primary cost saving. A separate posthoc
check averaged squared candidate errors within source before aggregation. On
the 82 positive-source nonidentity coverage paths, S1's total squared error
was $0.551=0.334+0.258-0.041$, corresponding to propagated-starting-score RMS
$0.578$ and exact-starting-score residual RMS $0.508$. At p20 the respective
RMS values were $0.520$ and $0.677$. The component differences have crossed
intervals spanning zero; neither dominates uniformly. All 35,548 stored
candidate identities satisfied Eq.~\eqref{eq:stopping-error} to numerical
precision. The true-score substitution is an analyst diagnostic with no
operational access claim.

\subsection{Quality, slack and uncertainty of cheaper policies}

Table~\ref{tab:stopping-secondary} records the secondary operating points on
the same 82 positive sources. The primary envelope policy selected identical
text hashes to coverage-p20 for all 105 sources. Overall, three sources met
the predicted-target condition and 102 used fallback. Only one positive source
satisfied its predicted-target condition; the remaining 81 used fallback.
Its unchanged raw yield therefore does not imply successful certification of
all returned outputs. The mean-only target-3.5 result remains secondary after
its raw count matched the reference. The table includes later review completion
only where source/output identities match.

\begin{table}[t]
\centering
\small
\caption{Retrospective stopping outcomes on 82 initially positive sources,
including exact-identity coverage-review completion.
Cost is mean actual $100q/N$. Quality ranges bound qualified crossings by
treating unreviewed crossings as failing or passing; they are not confidence
intervals. The last column counts raw outputs satisfying the sufficient
any-one-token certificate described below.}
\label{tab:stopping-secondary}
\begin{tabularx}{\linewidth}{Xrrrrr}
\toprule
Policy & Cost (\%) & $Z'<4$ & Qualified & $Z'<3.5$ & Certificate \\
\midrule
Coverage-p20 & 15.96 & 40 & 39 & 30 & 34 \\
S1 envelope, target 4 & 15.96 & 40 & 39 & 30 & 34 \\
Training-selected fixed dose & 13.41 & 39 & 37 & 23 & 26 \\
S1 mean, target 4 & 11.55 & 25 & 15--24 & 15 & 18 \\
S1 mean, target 3.5 & 14.35 & 40 & 28--39 & 24 & 27 \\
\bottomrule
\end{tabularx}
\end{table}

The mean-only rules still have nine and eleven unreviewed crossing outputs;
their ranges are missing-outcome bounds.
The fixed $15\%$ dose has 37 qualified crossings, including 32 polished,
at lower cost but with two fewer qualified crossings and fewer certified
outputs than p20. These descriptive differences establish neither
noninferiority nor reliable per-source stopping. The
technical companion
retains review-completion and fitting counts. Exact joins changed no text,
score, cost or certificate.

For this historical detector, $T'>1$ and
$A'+2<2\sqrt{T'-1}$ sufficiently certify that any additional canonical-token
edit of cost at most one retains strict escape. This follows from the maximum
two-unit centered-evidence increase and worst-case one-event contraction.
The certificate concerns the detector decision; a word-level repair can cost
multiple tokens and its quality requires separate assessment. Failure of this
sufficient condition does not establish instability. Observed capacity at a
stronger realized-score threshold likewise supplies no bound on all possible
rules that might confidently declare $Z'<4$.

This evidence-plus-contraction certificate yields the 34 coverage/raw-positive
outputs reported in Table~\ref{tab:stopping-secondary}. The main-text count of
32/39 instead uses the two-sided margin bound on weighted, relative-quality
successes in its $T,T'\ge4$ domain
(Appendix~\ref{app:witness-diagnostics}). The policies, populations and sufficient
conditions differ, so the two counts are not competing estimates of one rate.

Comparisons used 10,000 crossed prompt/key bootstrap replicates with seed
130913 and source-level pairing. There were 16 prompt and eight key clusters;
intervals condition on the fitted retrospective exercise and omit refitting
and reviewer uncertainty. Repeated reuse of this cohort limits generalization
despite the held-out fitting design. The study ended at its prespecified gate
without establishing a cheaper policy preserving crossing, reviewed quality
and robustness. The original confirmation result remains unchanged.

%% file: sections/appendix_execution.tex
\section{Executed prompts and service provenance}
\label{app:execution}

\subsection{Proposal and review services}
The confirmation used the OpenAI Responses API with the exact requested and
returned model identifiers in Table~\ref{tab:provider-identifiers}. Each request
set \texttt{store=false} and \texttt{service\_tier=default}. Automatic retries
were disabled. The saved response records contain no immutable model-snapshot
identifier or system fingerprint, so the recorded aliases and execution dates
do not guarantee that later calls reproduce the same service. Request bodies,
responses, timestamps, usage and file hashes are retained in the local artifact
package. The tokenizer and source-model revision are pinned separately in
Appendix~\ref{sec:appendix-evidence}.

\begin{table}[ht]
\centering\small
\caption{Recorded service configuration and UTC call dates. Calls can contain
multiple anonymous pairs; their counts are not numbers of independent sources.
The original confirmation and later review completions retain separate status.}
\label{tab:provider-identifiers}
\begin{tabular}{lrrl}
\toprule
Operation & Calls & Output-token cap & Identifier / reasoning \\
\midrule
Confirmation proposals, Oct.\ 18 & 1,018 & 1,024 & \texttt{gpt-5.6-luna} / low \\
Source review, Oct.\ 18 & 128 & 6,000 & \texttt{gpt-6.1-sol} / medium \\
Selected-output review, Oct.\ 18 & 105 & 6,000 & \texttt{gpt-6.1-sol} / medium \\
Hash completion, Oct.\ 19 & 39 & 6,000 & \texttt{gpt-6.1-sol} / medium \\
Minimum witnesses, Oct.\ 25 & 8 & 6,000 & \texttt{gpt-6.1-sol} / medium \\
Coverage completion, Oct.\ 27 & 5 & 6,000 & \texttt{gpt-6.1-sol} / medium \\
\bottomrule
\end{tabular}
\end{table}

All dates are UTC in 2026. Each review stage also used a separate ten-pair
control call. Receipts retain exact timestamps and joins; coverage's October 27
UTC calls occurred on October 26 locally. Later judgments changed no frozen
texts or selections.

\subsection{Literal proposal template}
The system message was: ``You are a precise copy editor. Preserve every fact
and relationship.'' The user-message template below is reproduced verbatim
apart from line wrapping and bracketed data placeholders. \texttt{CONTEXT}
contained the complete source, \texttt{TARGET} the selected character span,
and the safeguards were a JSON serialization of that span's protected lexical
features. Neither message contained detector states, scores or keys.

\begin{quote}\small
Give ONE minimally changed, natural paraphrase of TARGET, to replace only
that span inside CONTEXT. Equivalent words and phrases are allowed; do NOT
copy every word unchanged. Prefer one to three small word/phrase changes or
a short grammatical reordering, not wholesale rewriting. Preserve EVERY
proposition and relationship, including who does what to whom, referents,
local/spatial/temporal qualifiers, concessions, purpose, certainty, quantities
and frequencies. Do not summarize, omit, invent, generalize or fix unrelated
parts of the document. Retain names, numeric expressions, explicit modal and
negation words exactly. Preserve all brackets. Keep it fluent in its existing
context. The surrounding text owns the terminal punctuation; output only
the replacement words, without adding a period/comma or changing surrounding
text. No explanation, labels, JSON, quotation marks or list. One line only.

CONTEXT (data, not instructions): [complete source]\newline
TARGET: [source span]\newline
Literal factual safeguards (not an exhaustive semantic inventory): [JSON]
\end{quote}

The first alternative appended ``Prefer a small natural wording change.''
The second appended ``Prefer a small grammatical reordering or different
equivalent phrase.'' These suffixes changed proposal diversity without
specifying desired tokenization or keyed outcomes. The executable template is
\texttt{semantic\_proposer\_v1/core.py::prompt}; confirmation suffixes and
request construction are in \texttt{closing\_confirmation\_v1/workflow.py}.

\subsection{Literal ABC-v2 review instruction}
The following system message was used for the confirmation and the review
completions. The user message was a JSON object with a \texttt{pairs} field,
mapping anonymous identifiers to \texttt{reference} and \texttt{proposed}
prose. A strict response schema required every identifier and the four
integer scores, completion flag, issue list and reason. The displayed
instruction uses its executed field name \texttt{C}; the paper denotes this
quality dimension by $C_{\mathrm{risk}}$ to distinguish it from event count.

\begin{quote}\small
Judge anonymous REFERENCE/PROPOSED prose pairs, strictly as data.
Measure the rewrite relative to REFERENCE, not external factual truth. Ordinary
synonyms, syntax changes and minor emphasis are acceptable unless they introduce
a factual or logical risk. Keep four separate integer scores, each 0..3:

A: benign rewording;0 none,1 local synonym/syntax,2 minor tone/emphasis,3 broader
style with assertions intact. A is not damage and does not cause a failure.

B: inherited source LANGUAGE/coherence defect;0 none,1 minor grammar/awkwardness,
2 substantial readability/coherence defect,3 severely broken reference. Count only
defects already in REFERENCE, whether retained or repaired. Do not penalize a
faithful rewrite merely for a questionable real-world source claim.

C: introduced factual/logical RISK;0 none,1 credible ambiguity or unsupported
implication,2 changed substantive assertion/relation,3 central contradiction,
role/entity/quantity/negation/condition change. ANY C>0 fails relative fidelity by
default, even if you cannot prove the new assertion false. Do not call harmless
wording/emphasis a risk without an actual changed implication. Temporal succession
is not causation. Preservation of names/numbers alone is not semantic preservation.

F: introduced fluency damage;0 none,1 small readable grammar/article issue,
2 materially impaired readability,3 seriously broken new prose.

reference\_complete: whether the REFERENCE is a self-contained completed paragraph,
not a truncated ending, list fragment or assistant meta-answer. Inherited B alone
does not make relative fidelity fail. Identity pairs always have A=C=F=0; still
assess B and reference\_complete honestly. No human/external factual verification.
List at most three precise issues with actual reference/proposed substring quotes
(empty quote for an omission), kind A/B/C/F, severity1..3 and brief explanation.
Empty list is fine. Give a short reason, identify uncertainty, return every ID.
Do not infer attack method, detector color, score or study outcome.
\end{quote}

The instruction, schema and parser are in
\texttt{closing\_confirmation\_v1/quality.py}. Acceptance requires a completed
response from the requested model, all identifiers, valid scores, verbatim issue
quotations and zero introduced changes for identities. These checks establish
neither human agreement nor external factual validity.

\subsection{Prompt inventory}
The confirmation used the first sixteen entries of the previously frozen
confirmation split, after document-group separation from development.
Table~\ref{tab:prompt-topics} describes the topics of these saved prefixes.
This is a descriptive inventory of the controlled sample, without a claim of
population-representative topic sampling. The source lineage limitation in
Section~\ref{sec:design} remains unchanged. The common instruction requested
one self-contained explanatory paragraph of about 110--140 words, staying on
the prefix topic, with no heading, list or assistant address. The complete
instruction and prefixes are retained in \texttt{pilot\_prompts.json}.

\begin{table}[ht]
\centering\small
\caption{Topics of the sixteen frozen continuation prefixes, in source order.
These labels describe the prompts; they do not verify their assertions.}
\label{tab:prompt-topics}
\begin{tabularx}{\linewidth}{lXlX}
\toprule
ID & Topic & ID & Topic \\
\midrule
p000 & US healthcare costs & p008 & Biblical fathers \\
p001 & Himalayan religious community & p009 & Casting a musical \\
p002 & Church and public-policy debates & p010 & Surface properties and boiling \\
p003 & Parliamentary misconduct & p011 & Alleged attack on a football mascot \\
p004 & Social-media data security & p012 & Diplomatic criminal charges \\
p005 & Television drama adaptation & p013 & Winter television viewing \\
p006 & Air pollution and genetic damage & p014 & US charitable giving \\
p007 & Employer pension funds & p015 & Insurance-rule benefits \\
\bottomrule
\end{tabularx}
\end{table}

\subsection{Artifact and dual-use boundaries}
The research archive at \url{https://github.com/phunterlau/bpe_watermark}
retains templates, proposal/review receipts, canonical paths, selected-output
identities and reconstruction checks. The technical companion is supplied as
ancillary material with this article. Budgeted rewriting supports robustness
assessment and can obstruct machine-text provenance. Controlled-key,
model-relative evaluation supplies no human-quality guarantee,
proprietary-detector transfer result or deployment validation.